\documentclass[twoside]{article}

\usepackage[accepted]{aistats2024}
\usepackage[sort,round]{natbib}
\newcommand{\bH}{\bm{H}}
\newcommand{\bF}{\bm{F}}

\newcommand{\algoname}{\texttt{FedFisher}}
\newcommand{\algonamed}{\texttt{FedFisher(Diag)}}
\newcommand{\algonamek}{\texttt{FedFisher(K-FAC)}}

\newcommand{\prob}{\mathbb{P}}

\newcommand{\bA}{{\bm A}}
\newcommand{\bD}{{\bm D}}
\newcommand{\bT}{{\bm T}}
\newcommand{\bx}{{\bm x}}

\newcommand{\bb}{{\bm b}}

\newcommand{\bX}{{\bm X}}

\newcommand{\bU}{{\bm U}}
\newcommand{\bV}{{\bm V}}
\newcommand{\bSigma}{{\bm \Sigma}}

\newcommand{\by}{{\bm y}}

\newcommand{\bI}{{\bm I}}

\newcommand{\bw}{{\bm w}}

\newcommand\ind[1]{\mathbb{I}\left\{ #1 \right\}}
\newcommand\norm[1]{\left\| #1 \right\|_2}
\newcommand\ts[1]{\texttt{#1}}

\newcommand{\tmi}{\tilde{\bW}_i}
\newcommand{\omi}{\bW^*}
\newcommand{\osmi}{\bW^{\diamond}}

\newcommand{\tfi}{\bF_i}
\newcommand{\afi}{\tilde{\bF}_i}

\newcommand{\etas}{\tilde{\eta}_S}

\newcommand{\bW}{\bm{W}}

\newcommand{\ba}{\bm{a}}

\newcommand{\bB}{\bm{B}}
\newcommand{\bz}{\bm{z}}

\newcommand{\E}[1]{\mathbb{E}\left[{#1}\right]}
\newcommand{\Eg}[2]{\mathbb{E}_{#1}\left[{#2}\right]}
\newcommand{\tft}{\tilde{\bm{f}}}

\newcommand{\truey}{\tilde{\by}}

\newcommand{\floor}[1]{\lfloor {#1} \rfloor}

\newcommand{\wlmi}{\tilde{\bw}_{i,r}}
\newcommand{\wami}{\bw_{r}^{(0)}}
\newcommand{\womi}{\bw^*_{r}}

\newcommand{\wimi}{\bw_{0,r}}
\newcommand\bigO[1]{\mathcal{O}\left(#1 \right)}

\usepackage{hyperref}
\usepackage{url}
\usepackage{amsmath,amsfonts,bm}
\usepackage{amssymb}
\usepackage{amsmath}
\usepackage[utf8]{inputenc} 
\usepackage[T1]{fontenc}    

\usepackage{booktabs}       
\usepackage{wrapfig,lipsum,booktabs}
\usepackage{bm} 
\usepackage{nicefrac}       
\usepackage{microtype}      
\usepackage{xcolor}         
\usepackage{algorithm}
\usepackage{algorithmic}
\usepackage{amsthm}
\usepackage{titletoc}
\usepackage{subfig}
\usepackage{cleveref}
\usepackage{hyperref}
\usepackage{multirow}
\usepackage{makecell}
\usepackage{mathtools}
\usepackage{adjustbox}
\usepackage{booktabs}
\usepackage[font=small]{caption}
\usepackage{romannum}

\newtheorem{lem}{Lemma}
\newtheorem{thm}{Theorem}
\newtheorem{defn}{Definition}
\newtheorem{coro}{Corollary}
\newtheorem{clm}{Claim}

\newtheorem{prop}{Proposition}

\newtheorem{rem}{Remark}
\newtheorem{assum}{Assumption}

\DeclareMathOperator*{\argmin}{arg\,min}
\DeclareMathOperator*{\argmax}{arg\,max}
\DeclarePairedDelimiter{\ceil}{\lceil}{\rceil}

\makeatletter
\def\thickhline{%
  \noalign{\ifnum0=`}\fi\hrule \@height \thickarrayrulewidth \futurelet
   \reserved@a\@xthickhline}
\def\@xthickhline{\ifx\reserved@a\thickhline
               \vskip\doublerulesep
               \vskip-\thickarrayrulewidth
             \fi
      \ifnum0=`{\fi}}
\makeatother

\newlength{\thickarrayrulewidth}
\def\HiLi{\leavevmode\rlap{\hbox to \hsize{\color{yellow!50}\leaders\hrule height .8\baselineskip depth .5ex\hfill}}}

\begin{document}

\pagenumbering{arabic}

\twocolumn[

\aistatstitle{FedFisher: Leveraging Fisher Information for
One-Shot Federated Learning}

\aistatsauthor{ Divyansh Jhunjhunwala \And Shiqiang Wang \And  Gauri Joshi }

\aistatsaddress{ Carnegie Mellon University \And  IBM Research \And Carnegie Mellon University } ]

\begin{abstract}
 Standard federated learning (FL) algorithms typically require multiple rounds of communication between the server and the clients, which has several drawbacks, including requiring constant network connectivity, repeated investment of computational resources, and susceptibility to privacy attacks. One-Shot FL is a new paradigm that aims to address this challenge by enabling the server to train a global model in a single round of communication. In this work, we present \ts{\algoname}, a novel algorithm for one-shot FL that makes use of Fisher information matrices computed on local client models, motivated by a Bayesian perspective of FL. First, we theoretically analyze \ts{\algoname} for two-layer over-parameterized ReLU neural networks and show that the error of our one-shot \ts{FedFisher} global model becomes vanishingly small as the width of the neural networks and amount of local training at clients increases. Next, we propose practical variants of \ts{\algoname} using the diagonal Fisher and K-FAC approximation for the full Fisher and highlight their communication and compute efficiency for FL. Finally, we conduct extensive experiments on various datasets, which show that these variants of \ts{\algoname} consistently improve over competing baselines. 
\end{abstract}

\section{Introduction}

Data collection and storage are becoming increasingly decentralized, both due to the proliferation
of smart devices and privacy concerns stemming from transferring and storing data
at a centralized location. Federated Learning (FL) is a framework designed
to learn the parameters $\bW \in \mathbb{R}^d$ of a model $f(\bW,\cdot)$ on decentralized data distributed across a network of clients
under the supervision of a central server \citep{kairouz2019advances,li2020federated,yang2019federated}. 
Formally, we formulate FL as the following distributed optimization problem:
\begin{align}
    \min_{\bW \in \mathbb{R}^d} \left\{L(\bW) := \frac{1}{M}\sum_{i=1}^M L_i(\bW) \right\}\text{ where } \nonumber\\ L_i(\bW) = \frac{1}{n}\sum_{(\bx_{ij},\by_{ij}) \in \mathcal{D}_i} \ell(f(\bW,\bx_{ij}),\by_{ij}).
\label{eq:prob_form}
\end{align}%

Here, $M$ is the number of clients, $\mathcal{D}_i$ is the $i$-th client's local dataset consisting of
input-label pairs $\{(\bx_{ij},\by_{ij})\}_{j=1}^{n}$ where $\bx \in \mathbb{R}^p$ is the input
and $\by \in \mathbb{R}^C$ is the label, and $n = |\mathcal{D}_i|$ is the dataset size. For simplicity, we consider the case where clients have equal amounts of data; our algorithm and analysis can easily be extended to the case where client objectives are unequally weighted based on their local dataset sizes. The loss function $\ell(\cdot,\cdot)$ penalizes the difference between the
prediction of the model $f(\bW,\bx)$ and the true label $\by$. We use $\mathcal{D} = \{\mathcal{D}_i\}_{i=1}^M$ to denote the collection of data across all clients and $N = Mn$ to denote the total number of data samples across clients. 

The fundamental challenge with learning in federated settings is data heterogeneity across clients \citep{karimireddy2019scaffold,li2020federated}.
In particular, while clients can independently train a model that fits their local data, 
it is unclear how the server can combine these local models to get a global model that works for all the clients.
To tackle this problem, standard FL algorithms such as \texttt{FedAvg} \citep{mcmahan2017communication} and 
\texttt{FedProx} \citep{sahu2019federated} resort to an iterative approach requiring multiple rounds
of communication between clients and the server in order for the local models to reach a consensus. 
However, this multi-round approach has several drawbacks. First, clients need to frequently connect with the server to send and receive updates. 
Second, a client needs to invest computational resources repeatedly to update the global model every time it participates in training. 
Third, it increases susceptibility to attacks
such as data and model poisoning since attackers can continuously modify
and update their attacks based on the global model they receive from the server in each round. 

To overcome these drawbacks, a recent line of work has focused on the paradigm of \textit{one-shot FL} which aims to learn the parameters of the global model in a \textit{single} round of communication between clients and the server. Existing work for one-shot FL can be broadly split into two categories: (i) knowledge distillation methods and (ii) neuron matching methods.
Knowledge distillation methods treat the collection of client models as an ensemble and propose to distill the knowledge from this ensemble into a single global model. To perform the distillation step, some works assume that the server has access to an auxiliary public dataset \citep{lin2020ensemble,gong2021ensemble,li2020practical}, which clearly simplifies the problem.
Another set of works proposes training generative models such as GANs \citep{zhang2022dense,zhu2021data} or variational autoencoders \citep{heinbaughdata} to artificially generate data using local models of the clients. This again raises privacy questions, since a sufficiently powerful generative model can simply reconstruct client data \citep{hitaj2017deep}. Moreover, both the data generation and distillation steps impose a significant computation cost on the server and require careful hyperparameter tuning \citep{kurach2019large}, which is itself a challenge to implement in one-shot FL.

Neuron matching methods are based on the observation that neural networks (NNs) are permutation invariant, i.e., it is possible to create NNs that differ only in the ordering of weights while having the same output \citep{singh2020model, ainsworth2022git, akash2022wasserstein, liu2022deep, jordan2022repair, entezari2021role}. Based on this observation, these works propose to first align the weights of the client models according to a common ordering (called matching) and then average the aligned client models. Although this idea has been shown to work well when combining simple models like feedforward NNs, the performance drops considerably for more complex models such as CNNs \citep{wang2020federated}. Another line of work, termed as \textit{model fusion} has looked at fusing the capabilities of multiple existing models into a single model \citep{matena2022merging, choshen2022fusing, jin2022dataless, yadav2023resolving, ilharco2022editing}. While not explicitly designed for one-shot FL, some of these techniques, such as \citet{matena2022merging, jin2022dataless, yadav2023resolving} can be considered as an improvement over vanilla averaging. However, we note that these methods focus mostly on fusing pre-trained models and do not consider the effect of data heterogeneity used in training the models. We defer a more extensive discussion of such methods and other related work to \Cref{sec:appendix_related_work}. Finally, we note that none of these existing works provide any theoretical guarantees for their proposed methods. 

Thus, motivated by the limitations of multi-round FL and current approaches for one-shot FL, we ask the following questions: \textit{Can we devise a one-shot FL method that is simultaneously communication and computation efficient (at both the server and clients), privacy-preserving, and has good practical performance? Furthermore, can we provide theoretical guarantees for such a method?}

\paragraph{Our Contributions.} In this work, we take a step towards providing an affirmative answer to both of the questions formulated above. To do so, we use the idea that \Cref{eq:prob_form} can alternatively be reformulated as a \textit{posterior inference} problem, specifically finding the mode of a global posterior over the model parameters \citep{al2020federated, guo2023federated}. While \cite{al2020federated} and \cite{guo2023federated} use this idea to propose a multi-round algorithm, our contribution lies in showing how this reformulation can yield a novel one-shot FL algorithm, which we term as \ts{\algoname}. Some highlights of our contribution are as follows.
\begin{itemize}
    \item We propose \ts{\algoname} and show how the problem of finding the mode of a global posterior can be solved in a one-shot manner, using the local models at clients and some approximation of the Fisher information matrices computed at these local models (\Cref{sec:propose_algo}). 

    \item We theoretically analyze \ts{\algoname} for the case of over-parameterized two-layer neural networks. In particular, we show that when we utilize the full Fisher information in \ts{\algoname}, the error of our one-shot global model becomes vanishingly small as the width of the neural networks and the amount of local training at the clients increase (\Cref{sec:theoretical_analysis}). 

    \item We propose practical variants of \ts{\algoname} using the diagonal and K-FAC approximation, which we term as $\ts{\algonamed}$ and $\ts{\algonamek}$, respectively, and highlight the communication and compute efficiency of these variants along with their compatibility with secure aggregation (\Cref{sec:practical_algo}). 
    
    \item We evaluate \ts{\algonamed} and \ts{\algonamek} on a range of one-shot FL tasks using deep neural networks and show that they give a consistent $5-10$\% improvement in global model accuracy compared to competing one-shot baselines involving knowledge distillation, neuron matching, or model fusion  (\Cref{sec:experiments}). 
\end{itemize}

\section{Proposed Algorithm: FedFisher}

\label{sec:propose_algo}

To begin our discussion, we first state the following standard assumption on the loss function $\ell(\cdot,\cdot)$, which is true for most common loss functions such as the squared loss and the cross-entropy loss.

\begin{assum}
\label{assum:1}
Given $\bz = f(\bW,\bx)$, we assume that $\ell(\bz,\by)$ is proportional to the negative log likelihood of $\by$ under some exponential-family probabilistic model, i.e., $\ell(\bz,\by) \propto -\log \prob (\by|\bz)$ where $\prob(\by|\bz) = h(\by)\exp(\bz^{\top}T(\by)-A(\bz))$ and $h(\by),T(\by),A(\bz)$ are some real-valued functions.  
\end{assum}

Let us define the likelihood for a data point $(\bx,\by)$ for a given $\bW$ as $\prob((\bx,\by)|\bW) = \prob(\bx)\prob(\by|\bx, \bW) \propto q(\bx)\exp(-\ell(f(\bW,\bx_{ij}),\by))$, where $q(\cdot)$ is some prior on $\bx$, independent of $\bW$. Given this definition, we can adopt a Bayesian viewpoint and try to find the \textit{maximum a posteriori probability} (MAP) estimate, i.e., find $\bW$ where the \textit{posterior} likelihood $\prob(\bW|\mathcal{D}) \propto \prob(\mathcal{D}|\bW)\prob(\bW)$ is maximized with $\prob(\bW)$ being some prior belief over $\bW$. Our
motivation to do so comes from the following proposition. 

\begin{prop}\label{prop:posterior_decomp} (Global Posterior Decomposition \citep{al2020federated})
Under the flat prior $\prob(\bW) \propto 1$, the global posterior decomposes into a product of local posteriors, i.e., $\prob(\bW|\mathcal{D}) \propto \prod_{i=1}^M \prob(\bW|\mathcal{D}_i)$. Furthermore, the modes of the global posterior coincide with the optima of the FL objective in \Cref{eq:prob_form}, i.e, $\argmax_{\bW \in \mathbb{R}^d} \prob(\bW|\mathcal{D}) = \argmin_{\bW \in \mathbb{R}^d} L(\bW)$. 
\end{prop}

\Cref{prop:posterior_decomp} tells us that as long as clients compute and send their local posteriors $\prob (\bW|\mathcal{D}_i)$ to the server, no further server-client communication is needed to find the global MAP estimate or equivalently a minimizer to our FL objective, giving us a one-shot inference procedure. However, doing so is challenging since $\prob(\bW|\mathcal{D}_i)$ typically does not have an analytical expression. To get a tractable solution, we propose to use some approximate inference techniques, as discussed below. 

\paragraph{Mode of Local Posterior.}
To apply the approximate inference techniques detailed below, clients first need to compute $\tmi \approx \argmax_{\bW} \prob (\bW|\mathcal{D}_i)$, an estimate of the mode of their local posterior under the flat prior. Note that this corresponds to a minimizer of $L_i(\bW)$ under \Cref{assum:1} and therefore $\tmi$ can be obtained using standard gradient-based optimizers.

\paragraph{Laplace Approximation for Local Posterior.} 

Now using a second order Taylor expansion around $\tmi$, we can get the following approximation for the log-posterior at the $i$-th client: 
\begin{align}
\textstyle
    \log \prob(\bW|\mathcal{D}_i) 
    & \approx \log \prob(\tmi|\mathcal{D}_i) \nonumber\\
    & \hspace{10pt} - \frac{n}{2}(\bW - \tmi)^{\top}\bH_i(\bW - \tmi),\label{eq:laplace_taylor_exp}
\end{align}
where we additionally use $\nabla \log \prob(\bW|\mathcal{D}_i) |_{\bW = \tmi} \approx 0$. Here $\bH_i = - \frac{1}{n}\nabla^2 \log \mathbb{P}(\bW|\mathcal{D}_i)|_{\bW = \tmi}$ is the Hessian of the negative log-posterior at client $i$ computed at $\tmi$. Thus, we see that the local posterior of a client is now parameterized by $\tmi$ and $\bH_i$.

\paragraph{Approximating Hessian with Fisher.}  The Fisher information matrix $\bF_i$ (for brevity, we refer to it as ``the Fisher'' in the rest of the paper) of the local model $\tmi$ at client $i$ is defined as follows:
\begin{align}
\textstyle
    \bF_i &= \frac{1}{n}\sum_{j=1}^n \mathbb{E}_{\by} \left[ \nabla \log \prob(\by|\bx_{ij},\bW) \nabla \log \prob(\by|\bx_{ij},\bW) ^{\top} \right]
    \label{eq:fisher_decomp}
\end{align}
computed at $\bW = \tmi$. Now, under \Cref{assum:1} and assuming $\tmi$ fits the data at client $i$ perfectly, i.e., $f(\tmi,\bx_{ij}) = \by_{ij}, \forall j \in [n]$, it can be shown that the Hessian at $\tmi$ corresponds exactly to the Fisher, i.e., $\bH_i = \bF_i$ \citep{martens2020new, singh2020woodfisher}. The latter condition usually holds true for modern over-parameterized deep learning models when trained for sufficient epochs. Motivated by this observation, we can further approximate $\bH_i$ with $\bF_i$. This approximation is useful since, unlike the Hessian, the Fisher is guaranteed to be positive semi-definite, a condition required for tractable inference at the global server. However, computing (and communicating) the full Fisher would require $\mathcal{O}(d^2)$ bits, which is infeasible when the models are neural networks with $d$ in the order of millions. In practice, clients can replace the full Fisher $\bF_i$ with another computationally tractable approximation $\afi$ such as the diagonal Fisher or K-FAC \citep{grosse2016kronecker}, which preserves the positive semi-definiteness of $\bF_i$ (see discussion in \Cref{sec:practical_algo}). Thus we have
\begin{align}
\textstyle
\label{eq:hessian_approx}
   \bH_i \approx \tfi \approx \afi.
\end{align}
\paragraph{Computing Mode of Global Posterior.} 
Now assuming clients compute and send back $\tmi$ and $\afi$ to the server, we can use \Cref{prop:posterior_decomp}, \Cref{eq:laplace_taylor_exp} and \Cref{eq:hessian_approx} to approximate the logarithm of the global posterior as
\begin{align}
\textstyle
   \log \prob(\bW|\mathcal{D}) 
   & \approx \sum_{i=1}^M \log \prob(\tmi|\mathcal{D}_i) \nonumber\\
   & \hspace{10pt} - \frac{n}{2}\sum_{i=1}^M  (\bW-\tmi)^{\top}\afi(\bW-\tmi).
\label{eq:approx_global_post}
\end{align}

With this approximation, finding a mode of the global posterior can be written as the following optimization problem:
\begin{align}
\label{eq:argmax_approx_global_post}
    \min_{\bW \in \mathbb{R}^d}\,\, \sum_{i=1}^M (\bW-\tmi)^{\top}\afi(\bW-\tmi)
    .
\end{align}
Since each $\afi$ is positive semi-definite, a global minimizer of \Cref{eq:argmax_approx_global_post} can be found by simply setting the derivative of the objective to zero. Doing so, we have the following proposition.
\begin{prop}\label{prop:posterior_solution}
Any $\bW$ that satisfies $(\sum_{i=1}^M \afi) \bW = \sum_{i=1}^M \afi \tmi$ is a minimizer of the objective $\sum_{i=1}^M (\bW - \tmi)^{\top}\afi(\bW-\tmi)$. 
\end{prop}

\setlength{\textfloatsep}{1em}
\begin{algorithm} [t]
\caption{\ts{\algoname} }\label{algo1}
\renewcommand{\algorithmicloop}{\textbf{Global server do:}}
\begin{algorithmic}[1]
\STATE {\bfseries Input:} initial $\bW_0$, no. of iterations $K$, $T$, client and server step sizes $\eta$ and $\eta_S$ respectively
\STATE {\bfseries Global server does:}\\
\STATE \hspace*{1em} Communicate $\bW_0$ to all clients; 
\STATE {\bfseries Clients $i \in [M]$ in parallel do:}
\STATE \hspace*{1em} Set $\bW_i^{(0)} \leftarrow \bW_0$;
\STATE \hspace*{1em} {\bfseries For ${k=0,\ldots,K-1}$ iterations:}\\
\STATE \hspace*{2em} $\bW_i^{(t+1)} \leftarrow \bW_i^{(t)} - \eta \nabla L_i(\bW_i^{(t)})$;
\STATE \hspace*{1em} {Set $\tmi \leftarrow \bW_i^{(K)}$;}
\STATE \hspace*{1em} {Compute $\afi$;} \hspace{5pt} \textcolor{gray}{// Approximation to true Fisher}

\STATE {\hspace*{1em} Communicate $\tmi$ and $\afi$ to the server;}
\STATE {\bfseries Global server does:}\\
\STATE \hspace*{1em} Set $\bW^{(0)} = \sum_{i=1}^M \tmi/M$;

\STATE \hspace*{1em} {\bfseries For ${t=0,\ldots,T-1}$ iterations:}\\
\STATE {\hspace*{2em}} $\bW^{(t+1)} \leftarrow \bW^{(t)} - \eta_{S}\sum_{i=1}^M \left(\afi\bW^{(t)} - \sum_{i=1}^M \afi\tmi \right)$;
\end{algorithmic} 
\end{algorithm}

For over-parameterized models, i.e, $d \gg N$, the rank of $\sum_{i=1}^M \afi$ will be smaller than $d$ and therefore the system of equations $(\sum_{i=1}^M \afi) \bW = \sum_{i=1}^M \tfi\tmi$ will not have a unique solution. To resolve this, we propose to use the solution that minimizes the sum of distances from the local models $\tmi$ of each client, i.e., minimizing $\sum_{i=1}^M \norm{\bW - \tmi}^2$. Such a constraint ensures fairness of the \ts{\algoname} objective towards all clients and prevents the solution from drifting too far away from the local models of each client. We express this mathematically as follows. 
\begin{align}
\textstyle
\label{eq:fisher_avg_sol}
    & \text{\ts{\algoname} objective: } \nonumber\\
    & \min_{\bW \in \mathbb{R}^d} \Bigg\{\ \tilde{L}(\bW) = \sum_{i=1}^M \norm{\bW - \tmi}^2, \nonumber \\ 
    &\quad\quad\quad\quad \text{ such that } \left(\sum_{i=1}^M \afi\right)\bW = \sum_{i=1}^M\afi \tmi \Bigg\}.
\end{align}

We refer to the minimizer $\omi$ of \Cref{eq:fisher_avg_sol} as the \ts{\algoname} global model in the rest of our discussion. We now show that $\bW^*$ can be easily computed using gradient descent (GD) with proper initialization and learning rate conditions via the following lemma.

\begin{lem}
Let $\bW^{(1)},\bW^{(2)},\dots$ be the iterates generated by running the following gradient descent (GD) procedure: $\bW^{(t+1)} = \bW^{(t)} - \eta_S \sum_{i=1}^M\left(\afi\bW^{(t)} - \afi\bW_i\right)$ with $\bW^{(0)} = \sum_{i=1}^M \tmi/M$ and $\eta_S \leq 1/\lambda_{\max}$ where $\lambda_{\max}$ is the maximum eigenvalue of $\sum_{i=1}^M \afi$. Then, $\lim_{T \rightarrow \infty} \bW^{(T)} = \bW^*$.
\label{lemma:lemma_gd_proj}
\end{lem}

Proofs for all lemmas and theorems in this paper can be found in \Cref{sec:appendix_proofs}. This concludes our discussion of the proposed algorithm \ts{\algoname}. An algorithm outline for \ts{\algoname} can be found in \Cref{algo1}, with practical implementation considerations discussed in \Cref{sec:practical_algo}. Next, we analyze the performance of \ts{\algoname} for over-parameterized two-layer networks.

\section{Theoretical Analysis for Two-layer Over-parameterized Neural Network}
\label{sec:theoretical_analysis}

In this section, we characterize the performance of the \ts{\algoname} global model $\bW^*$, on the true global objective $L(\bW)$ when $f(\bW,\cdot)$ is a two-layer over-parameterized neural network.

\paragraph{Sources of Error for \ts{\algoname}.}  We see that the mismatch between \ts{\algoname} objective $\tilde{L}(\bW)$ in \Cref{eq:fisher_avg_sol} and $L(\bW)$ can be characterized into the following sources of error: \textbf{(i)}  the suboptimality of $\tmi$, i.e, $\nabla L_i(\tmi) \neq \bm{0}$, 
\textbf{(ii)} the Laplace approximation, i.e., neglecting higher order terms in the Taylor expansion in
\Cref{eq:laplace_taylor_exp}, and \textbf{(iii)} the error introduced by approximating the true Fisher $\tfi$ with $\afi$ in \Cref{eq:hessian_approx}. The error introduced by $\textbf{(iii)}$ is orthogonal to $\textbf{(i)}$ and $\textbf{(ii)}$ and depends on the specific approximation used. Moreover, commonly used approximations, such as the diagonal and K-FAC, are primarily empirically motivated and have limited theoretical guarantees. Therefore, to simplify our analysis, we assume that clients send their full Fisher and focus on bounding the error due to the first two sources.

\paragraph{Novelty of Analysis.} 
 
 While the error due to \textbf{(i)} can be thought of as a function of the number of local training steps $K$, it is less clear how the error due to \textbf{(ii)} can be bounded. In this analysis, we show that for over-parameterized two-layer ReLU NNs, the error due to \textbf{(ii)} can be controlled using the \textit{width} of the network. Intuitively, the quality of the approximation in \Cref{eq:laplace_taylor_exp} depends on the distance between the weights of the \ts{\algoname} global model $\bW^*$ and clients' local models $\tmi$; the larger the distance the greater will be the effect of the higher-order terms and hence worse the approximation. Our contribution lies in showing that for a sufficiently wide model, this distance decreases as $\mathcal{O}(1/m)$ where $m$ is the width of the model. Doing so requires a careful analysis of the trajectory of iterates generated by the optimization procedure in \Cref{lemma:lemma_gd_proj} and novel bounds on the smallest and largest eigenvalues of the aggregated Fisher $\sum_{i=1}^M \tfi/M$ (see proofs in \Cref{sec:appendix_proofs}). We believe that our analysis can also be extended to deeper NNs as future work.

We begin by modeling a two-layer ReLU NN as follows:
\begin{align}
     f(\bW,\bx) = \frac{1}{\sqrt{m}}\sum_{r=1}^m a_r \sigma(\bx^{\top}\bw_r).
\end{align}
Here, $m$ is the number of hidden nodes in the first layer, $\{\bw_r\}_{r=1}^M$ are the weights of the first layer, $\{a_r\}_{r=1}^m$ are the weights of the second layer, and $\sigma(x) = \max\{x,0\}$ is the ReLU function. We use $\bW = \mathrm{vec}(\bw_1,\bw_2,\dots,\bw_m)$ to parameterize the weights of the first layer. Similar to \cite{du2018gradient}, we consider $a_r$'s to be fixed beforehand (initialized to be $+1$ or $-1$ uniformly at random) and only consider the case where $\bw_r$'s are trained. We also consider the squared loss function $\ell(z,y) = \frac{1}{2}(y-z)^2$.

Recall that $\mathcal{D}$ is the collection of data across all the clients. Define $\{(\bx_{k},y_{k})\}_{k=1}^N = \{(\bx_{11},y_{11}), (\bx_{12},y_{12}),\dots, (\bx_{Mn},y_{Mn})\}$ to be an ordering of the data in $\mathcal{D}$. We state some definitions and assumptions that will be needed in our analysis.

\begin{defn} (Minimum eigenvalue of Gram Matrix \citep{du2018gradient})
    Define the matrix $\bH^{\infty} \in \mathbb{R}^{N \times N}$ as $\bH_{k,l}^{\infty} = \mathbb{E}_{\bw \sim \mathcal{N}(\bm{0},\bm{I})}\left[\bx_{k}^{\top}\bx_{l}\mathbb{I}\left\{ \bw^{\top}\bx_{k} \geq 0\right\}\mathbb{I}\left\{\bw^{\top}\bx_{l} \geq 0\right\} \right]$ and $\lambda_0 = \lambda_{\min}(\bH^\infty)$. 
\label{defn:gram_matrix}
\end{defn}

\begin{assum} (Data normalization)
    For any $(\bx,y) \in \mathcal{D}$, we have $\norm{\bx} = 1$ and $|y| \leq C$, where $C$ is some positive constant.
\label{assum:data_normalization}
\end{assum}

\begin{assum} (Data uniqueness) For any $(\bx,y),(\bx',y') \in \mathcal{D}$, we have $\norm{\bx - \bx'} \geq \phi$, where $\phi$ is some strictly positive constant. 
\label{assum:data_separation}
\end{assum}

\begin{assum} (Full Fisher) Clients compute and send their full Fisher, i.e, $\afi = \tfi$. 
\label{assum:local_training} 
\end{assum}

\paragraph{Note on Definition and Assumptions.} 
The rate of convergence of GD for two-layer ReLU neural networks depends closely on $\bH^\infty$ and $\lambda_0$ as shown in \cite{du2018gradient}. \Cref{assum:data_normalization} and \Cref{assum:data_separation} are standard in NN optimization literature \citep{du2018gradient, zou2019improved, allen2019learning, allen2019convergence} and are needed to ensure bounded loss at initialization and $\lambda_0 > 0$, respectively. \Cref{assum:local_training} is specific to our setting, as discussed at the start of this section.

\begin{thm}
\label{theorem:fisher_avg_error}
Under Assumptions \ref{assum:data_normalization}, \ref{assum:data_separation}, \ref{assum:local_training}, for $m \geq \mathrm{poly}(N, \lambda_0^
{-1},\delta^{-1},\kappa^{-1})$, and i.i.d Gaussian initialization weights of $\bW_0$ as $\bw_{0,r} \sim \mathcal{N}(\bm{0},\kappa)$, and initializing the second layer weights $a_r = \{-1,1\}$ with probability $1/2$ for all $r \in [m]$, for step sizes $\eta = \mathcal{O}(\lambda_0/N^2)$, $\eta_S = \mathcal{O}(\lambda_0/N^2)$ and for a given failure probability $\delta \in (0,1)$, the following is true with probability $1-\delta$ over the random initialization:
\begin{align}
\label{thm:fisher_avg_error}
   L(\bW^*) & \leq  \underbrace{\mathcal{O}\left((1-\eta\lambda_0/2)^K \frac{N}{\delta}\right)}_{\textnormal{local optimization error}} \hspace{10pt} + \nonumber\\ 
   &  \underbrace{\mathcal{O}\!\left(\!(2-(1\!-\!\eta\lambda_0/2)^K)\frac{\mathrm{poly}(N, \kappa^{-1}, \lambda_0^
{-1},\delta^{-1})}{m}\right)}_{\textnormal{Laplace approximation error}} \! .
\end{align}
\end{thm}

\begin{coro}
\label{coro:fisher_avg_error}
Under the conditions of \Cref{theorem:fisher_avg_error}, if we set $m = \mathcal{O}\left(\frac{\mathrm{poly}(N, \kappa^{-1}, \lambda_0^
{-1},\delta^{-1})}{\epsilon}\right)$ and $K = \mathcal{O}\left(\frac{1}{\eta \lambda_0}\log (N/\delta\epsilon)\right)$, then for failure probability $\delta \in (0,1)$ and target error $\epsilon \in (0,1)$, the global loss $L(\bW^*) \leq \epsilon$ with probability $1-\delta$.
\end{coro}

\textbf{Takeaways.} \Cref{theorem:fisher_avg_error} shows that for sufficiently wide networks, $L(\bW^*)$ can be decomposed into two distinct sources of error. The first term is the local optimization error, which measures how well $\tmi$ fits the data at the $i$-th client, and the second term corresponds to the error introduced by the Laplace approximation in \Cref{eq:laplace_taylor_exp}. While the overall error always decreases as the width $m$ increases, there is a trade-off in the number of local optimization steps $K$. In particular, a larger $K$ reduces the local optimization error but increases the Laplace error as each local model $\tmi$ drifts further away from $\bW^*$.
\Cref{coro:fisher_avg_error} shows that by setting $m$ and $K$ large enough, the total training error can be made vanishingly small in just one round.

\paragraph{Effect of Data Heterogeneity.} Note that our result does not make any explicit assumptions on the heterogeneity of data across clients such as bounded dissimilarity \citep{wang2020tackling, karimireddy2019scaffold}.
We find that the notion of heterogeneity that implicitly appears in our analysis is the distance between the weights of any two local models, i.e, $\|\tilde{\bw}_{i,r} - \tilde{\bw}_{j,r}\|_2$ for any $i \neq j$. For two-layer neural networks this quantity can be bounded as follows: $\|\tilde{\bw}_{i,r} - \tilde{\bw}_{j,r}\|_2 \leq \|\tilde{\bw}_{i,r} - \bw_{0,r}\|_2 + \|\tilde{\bw}_{j,r} - \tilde{\bw}_{0,r}\|_2 \leq \mathcal{O}(1/\sqrt{m})$ (see first part of \Cref{lemma:global_opt} in \Cref{sec:appendix_proofs}), where $\bw_{0,r}$ are the weights of the common initialization point across clients. This bound is agnostic to the correlation of data across clients and shows that $\|\tilde{\bw}_{i,r} - \tilde{\bw}_{j,r}\|_2$ always decreases as the width of the model increases. Improving this bound by incorporating the effects of data correlation can be an interesting direction for future work.

\begin{figure}[h]
  \centering
   \subfloat[]{\includegraphics[width=0.5\linewidth]{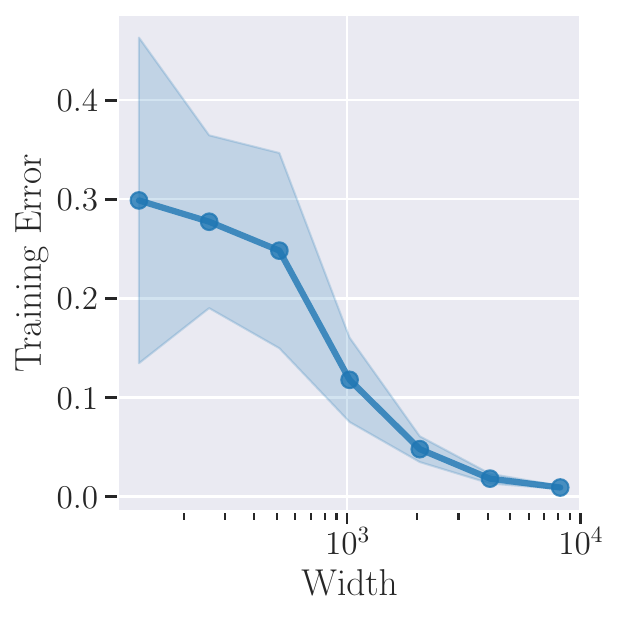}\label{fig:error_vs_width}}
   \subfloat[]{\includegraphics[width=0.5\columnwidth]{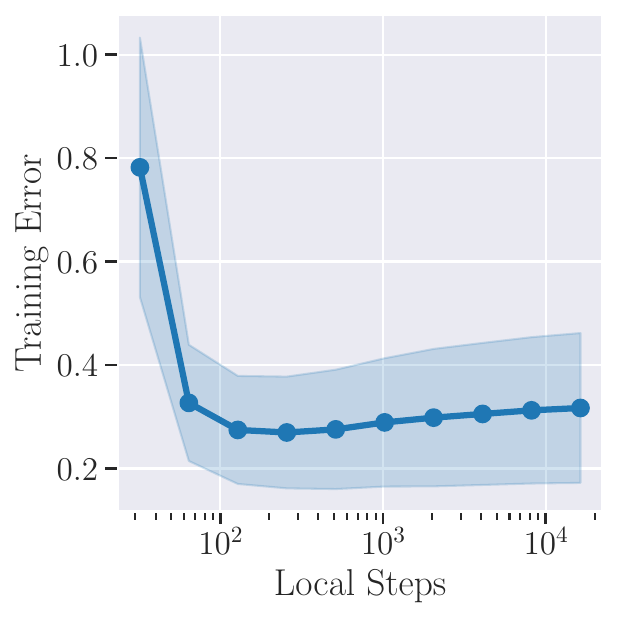}\label{fig:error_vs_local_steps}}\\
   \vspace{-0.5em}
   \caption{\small{Empirical validation of \Cref{theorem:fisher_avg_error} in a synthetic setting. For a fixed number of local steps (\Cref{fig:error_vs_width}), the error for \ts{\algoname} decreases as the width of the model increases. For a fixed width (\Cref{fig:error_vs_local_steps}), the error first decreases and then increases as local steps increases.}}
  \label{fig:toy_expts}
\end{figure}

\textbf{Empirical Validation.}  We conduct experiments on a synthetic dataset with $M =2$ clients, $p =2$ dimensions, and $n = 100$ samples to verify our theoretical findings. This setting allows us to ensure that the conditions in \Cref{theorem:fisher_avg_error}, such as computing the full Fisher, are computationally tractable. The synthetic data is generated following a similar procedure as \cite{sahu2019federated}. Further details of the experimental setup can be found in \Cref{sec:appendix_addnl_expts}. 
In \Cref{fig:error_vs_width} we fix the number of local steps $K = 2048$ and vary the width of the model. As predicted by our theory, we see that the training error for \ts{\algoname} monotonically decreases as we increase the width of the model. In \Cref{fig:error_vs_local_steps} we fix the width of the model to $m = 512$ and vary the number of local steps. Here we see a trade-off in the effect of local steps --  the error initially decreases as clients do more local steps and then rises again, as predicted by our theory. 

\paragraph{Generalization Bounds.} Motivated by \cite{arora2019fine}, we can also provide generalization bounds on the performance of the \ts{\algoname} global model by bounding $\norm{\bW^* - \bW_0}$, which is the \textit{total distance} between the weights of $\bW^*$ and the initialization point $\bW_0$. We provide these bounds and subsequent discussion in \Cref{sec:appendix_proofs}.

\section{A Practical Implementation of FedFisher}
\label{sec:practical_algo}

Two of the most popular approximations for the Fisher are the diagonal Fisher and the Kronecker Factored Approximate Curvature (K-FAC). The diagonal Fisher simply approximates the full Fisher with its diagonal elements and has been used in previous work for applications such as network pruning \citep{lecun1989optimal} and transfer learning \citep{kirkpatrick2017overcoming}. The K-FAC introduced by \cite{martens2015optimizing, grosse2016kronecker} uses two approximations: (i) it assumes that the Fisher is block diagonal where each block corresponds to the Fisher of a particular layer, and (ii) it approximates the Fisher of each block as the Kronecker product of two smaller matrices, i.e, $\tilde{\bF} \approx \bA \otimes \bB $. Thus, storing (and communicating) the K-FAC only takes $\mathcal{O}\left(\mathrm{dim}(\bA) + \mathrm{dim}(\bB)\right)$ bits instead of $\mathcal{O}\left(\mathrm{dim}(\bA)\cdot\mathrm{dim}(\bB)\right)$ bits. Using these approximations as a substitute for the true Fisher in \Cref{eq:hessian_approx}, we get two practical variants of \ts{\algoname}, which we term as \ts{\algonamed} and \ts{\algonamek}, respectively. We highlight the computation and communication efficiency of these variations along with their compatibility with secure aggregation below.

\paragraph{Computation Efficiency.} To compute their diagonal Fisher or K-FAC, clients need to perform an additional forward-backward pass over the data, i.e., an additional epoch of training, plus some small overhead cost. This is a reasonable cost for FL setups, since most of the computation cost goes into computing $\tmi$, which requires multiple local epochs. This is demonstrated in \Cref{table:compute_time}, where we see that \ts{\algonamed} and \ts{\algonamek} add less than $14.5\%$ of the total computational cost of \ts{FedAvg} at clients for our FL setups described in \Cref{sec:experiments}.

\begin{table}[h]
\centering
\caption{\small{Computation time in seconds averaged across $M = 5$ clients for FL setup described in \Cref{sec:experiments}.}}
\small
\color{black}

\begin{tabular}{lcc|cc}
\toprule
\multicolumn{1}{l@{\hspace{-50pt}}}{Method} &
\multicolumn{2}{c}{FashionMNIST} &
\multicolumn{2}{c}{CIFAR10}\\
\midrule
\texttt{FedAvg} 
& $14.5$ & $-$
& $20.9$ & $-$\\

\texttt{\algonamek} 
& $16.0$ & $+10.3\%$
& $23.1$ & $+10.5\%$\\

\texttt{\algonamed} 
& $16.6$ & $+14.4\%$
& $23.7$ & $+13.4\%$\\

\bottomrule
\end{tabular}
\label{table:compute_time}
\end{table}

\begin{table*}[hbt!]
\centering
\caption{ \small Test accuracy results on different datasets by keeping number of client $M = 5$ fixed and varying heterogeneity parameter $\alpha$ (smaller $\alpha$ is more heterogeneous). \ts{\algoname} variants consistently outperform other baselines.}
\small
\label{table:heterogeneity}
{
\begin{tabular}{l@{\hspace{7pt}}c@{\hspace{6pt}}|@{\hspace{4pt}}c@{\hspace{7pt}}c@{\hspace{7pt}}c@{\hspace{7pt}}c@{\hspace{7pt}}c@{\hspace{7pt}}c@{\hspace{3pt}}|@{\hspace{4pt}}c@{\hspace{7pt}}c}\\
\toprule
Dataset & $\alpha$ & \texttt{FedAvg} & \texttt{OTFusion} & \texttt{PFNM} & \texttt{RegMean} & \texttt{DENSE} &  \texttt{Fisher} & \texttt{\algoname} & \texttt{\algoname}\\
 & & & & & & & \ts{Merge} & \ts{(Diag)} & \ts{(K-FAC)}\\
\midrule
& $0.2$ 
& $59.11$\tiny{$\pm 3.82$} 
& $58.40$\tiny{$\pm 4.95$} 
& $63.50$\tiny{$\pm 3.07$} 
& $71.09$\tiny{$\pm 2.19$} 
& $72.69$\tiny{$\pm 1.99$} 
& $61.81$\tiny{$\pm 3.98$} 
& $65.44$\tiny{$\pm 3.04$} 
& $\bf{76.28}$\tiny{$\pm 2.56$}\\ 

FashionMNIST 
& $0.1$ 
& $41.72$\tiny{$\pm 4.54$} 
& $43.77$\tiny{$\pm 2.03$} 
& $59.27$\tiny{$\pm 3.19$} 
& $56.98$\tiny{$\pm 4.08$} 
& $50.11$\tiny{$\pm 4.99$} 
& $54.71$\tiny{$\pm 4.96$} 
& $55.04$\tiny{$\pm 4.15$} 
& $\bm{68.36}$\tiny{$\pm 3.44$}\\

& $0.05$ 
& $36.02$\tiny{$\pm 2.77$} 
& $37.53$\tiny{$\pm 1.20$} 
& $45.69$\tiny{$\pm 5.32$} 
& $50.40$\tiny{$\pm 2.69$} 
& $46.62$\tiny{$\pm 3.21$} 
& $43.03$\tiny{$\pm 5.51$} 
& $45.92$\tiny{$\pm 6.13$} 
& $\bm{53.29}$\tiny{$\pm 4.17$}\\

\midrule

& $0.2$ 
& $63.39$\tiny{$\pm 2.27$}
& $68.05$\tiny{$\pm 0.98$} 
& $69.11$\tiny{$\pm 1.35$} 
& $74.04$\tiny{$\pm 1.20$} 
& $\bf{77.73}$\tiny{$\pm 2.07$} 
& $65.42$\tiny{$\pm 3.72$} 
& $68.68$\tiny{$\pm 3.42$} 
& $75.92$\tiny{$\pm 1.16$}\\

SVHN 
& $0.1$ 
& $39.42$\tiny{$\pm 2.52$} 
& $52.34$\tiny{$\pm 0.18$} 
& $53.19$\tiny{$\pm 4.18$} 
& $64.16$\tiny{$\pm 3.51$}
& $56.31$\tiny{$\pm 2.17$}
& $64.06$\tiny{$\pm 2.86$} 
& $64.20$\tiny{$\pm 3.08$} 
& $\bm{69.09}$\tiny{$\pm 2.81$}\\ 

& $0.05$
& $27.03$\tiny{$\pm 0.71$}
& $37.24$\tiny{$\pm 0.88$} 
& $45.62$\tiny{$\pm 3.55$} 
& $55.83$\tiny{$\pm 4.44$} 
& $49.16$\tiny{$\pm 1.63$} 
& $49.32$\tiny{$\pm 2.83$} 
& $51.48$\tiny{$\pm 2.91$} 
& $\bm{57.41}$\tiny{$\pm 3.79$}\\

\midrule

& $0.2$ 
& $41.77$\tiny{$\pm 0.79$} 
& $40.35$\tiny{$\pm 1.96$} 
& $45.75$\tiny{$\pm 0.58$} 
& $43.41$\tiny{$\pm 1.56$} 
& $44.79$\tiny{$\pm 1.04$} 
& $39.90$\tiny{$\pm 3.21$} 
& $44.04$\tiny{$\pm 2.10$} 
& $\bm{51.67}$\tiny{$\pm 1.03$}\\

CIFAR10 
& $0.1$ 
& $36.43$\tiny{$\pm 2.51$} 
& $40.45$\tiny{$\pm 0.96$} 
& $39.43$\tiny{$\pm 1.80$} 
& $36.65$\tiny{$\pm 1.29$} 
& $38.65$\tiny{$\pm 2.67$} 
& $36.53$\tiny{$\pm 2.74$} 
& $40.04$\tiny{$\pm 1.35$} 
& $\bm{47.01}$\tiny{$\pm 1.87$}\\

& $0.05$ 
& $29.75$\tiny{$\pm 1.32$} 
& $30.62$\tiny{$\pm 1.20$} 
& $30.58$\tiny{$\pm 3.65$} 
& $32.84$\tiny{$\pm 1.39$} 
& $32.65$\tiny{$\pm 2.35$} 
& $30.89$\tiny{$\pm 1.11$} 
& $31.08$\tiny{$\pm 1.52$} 
& $\bm{40.02}$\tiny{$\pm 3.64$}\\ 

\midrule
& $0.2$ 
& $37.15$\tiny{$\pm 0.87$} 
& $36.88$\tiny{$\pm 1.44$}
& $39.94$\tiny{$\pm 0.60$} 
& $36.19$\tiny{$\pm 1.07$}
& $35.64$\tiny{$\pm 1.22$} 
& $37.01$\tiny{$\pm 4.07$} 
& $41.88$\tiny{$\pm 2.53$} 
& $\bm{43.45}$\tiny{$\pm 0.54$}\\

CINIC10 
& $0.1$ 
& $32.43$\tiny{$\pm 1.57$} 
& $34.92$\tiny{$\pm 1.68$} 
& $35.86$\tiny{$\pm 1.65$} 
& $31.54$\tiny{$\pm 0.67$} 
& $30.89$\tiny{$\pm 1.44$} 
& $32.91$\tiny{$\pm 2.44$} 
& $37.68$\tiny{$\pm 2.21$}
& $\bm{39.16}$\tiny{$\pm 1.19$}\\

& $0.05$ 
& $27.58$\tiny{$\pm 0.76$} 
& $26.39$\tiny{$\pm 0.35$} 
& $29.04$\tiny{$\pm 1.26$} 
& $28.14$\tiny{$\pm 0.79$}
& $28.13$\tiny{$\pm 1.10$} 
& $29.77$\tiny{$\pm 1.05$} 
& $31.72$\tiny{$\pm 1.48$} 
& $\bm{32.90}$\tiny{$\pm 2.60$}\\ 
\bottomrule
\end{tabular} 
}
\end{table*}

\paragraph{Communication Efficiency.} The communication cost of \ts{\algoname} depends on the number of parameters in the approximate Fisher. For the diagonal Fisher this is exactly $d$, while for the K-FAC it is model specific, but usually less than $10d$. We find that the total communication cost of both \ts{\algonamed} and \ts{\algonamek} (including local models) can be easily reduced to match that of \ts{FedAvg} using a mix of standard compression techniques such as quantization and low-rank decomposition without significantly affecting the accuracy of the final one-shot model. We discuss more details about this compression in \Cref{sec:appendix_comm_eff}. For all our experiments in \Cref{sec:experiments}, we ensure that the communication cost of \ts{\algonamed} and \ts{\algonamek} matches that of \ts{FedAvg}.

\paragraph{Compatibility with Secure Aggregation.} In theory, the server optimization in \ts{\algoname} just needs the aggregate quantities $\sum_{i=1}^M \afi$, $\sum_{i=1}^M \afi\tmi$ and $\sum_{i=1}^M \tmi$ (see \Cref{algo1}). For the diagonal Fisher, these quantities can be computed using secure aggregation \citep{bonawitz2016practical,kadhe2020fastsecagg}, thus preventing the server from accessing individual $\tmi$'s and $\afi$'s and improving privacy. For K-FAC, however, while the server does not need access to individual $\tmi$'s, it does need access to individual $\afi$'s. This is because there does not exist a summation operation over K-FAC matrices that preserves the Kronecker factorization property. In other words, to compute matrix vector products of the form $\sum_{i=1}^M \afi \bx$, the server needs to store each $\afi$ and individually compute and sum up $\afi\bx$. However, we argue that having access to $\afi$ can be more private than having access to local models $\tmi$, where the latter is needed for knowledge distillation and neuron matching baselines. 
We discuss this in more detail, including an experiment on measuring privacy using model inversion attacks \citep{zhang2020secret}, in \Cref{sec:appendix_privacy_expt}.

\section{Experiments}
\label{sec:experiments}

\subsection{Setup}

\begin{figure*}[t]
    \centering
\includegraphics[width=2.00\columnwidth]{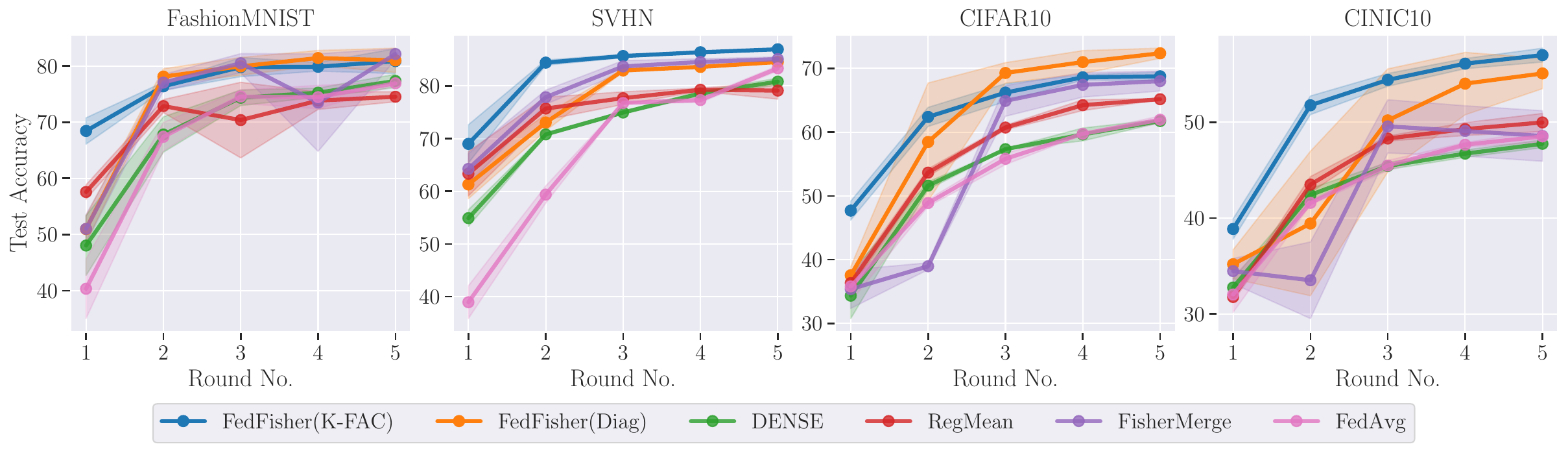}
    \caption{\small{Results of performing $5$ rounds of local training and aggregation across different datasets for $\alpha = 0.1$ and $M = 5$. \ts{\algoname} variants offer additional utility in multi-round settings and continue to improve over baselines.}}
    \label{fig:multi_round}
\end{figure*}

\setlength{\extrarowheight}{1.5pt} 
\begin{table*}[h]
\centering
\caption{ \small Results on one-shot aggregation using a pre-trained ResNet-18 model with $\alpha = 0.1$ and $M = 5$. \ts{\algoname} variants show a significant improvement in performance compared to baselines when using a pre-trained model.}
\small
\label{table:pretrained}
{
\begin{tabular}{c|ccccc|cc}
\toprule
Dataset 
& \texttt{FedAvg} 
& \texttt{OTFusion} 
& \texttt{RegMean} 
& \texttt{DENSE} 
& \texttt{Fisher} 
& \texttt{\algoname} 
& \texttt{\algoname}
\\
& & & & & \ts{Merge} & \ts{(Diag)} & \ts{(K-FAC)}\\
\midrule
CIFAR10
& $56.89$\tiny{$\pm 1.31$} 
& $56.89$\tiny{$\pm 1.33$} 
& $60.98$\tiny{$\pm 1.84$}  
& $57.18$\tiny{$\pm 1.11$} 
& $78.76$\tiny{$\pm 2.30$} 
& $79.06$\tiny{$\pm 0.82$} 
& $\bf{80.42}$\tiny{$\pm 1.13$}\\

CINIC10
& $50.05$\tiny{$\pm 0.70$} 
& $50.04$\tiny{$\pm 0.69$} 
& $50.53$\tiny{$\pm 0.69$}  
& $50.83$\tiny{$\pm 0.32$}
& $66.89$\tiny{$\pm 1.87$} 
& $\bm{69.33}$\tiny{$\pm 3.44$} 
& $68.04$\tiny{$\pm 1.64$}\\

GTSRB
& $40.63$\tiny{$\pm 3.30$} 
& $40.40$\tiny{$\pm 3.35$} 
& $46.05$\tiny{$\pm 2.52$} 
& $44.87$\tiny{$\pm 1.13$} 
& $61.27$\tiny{$\pm 2.80$} 
& $62.80$\tiny{$\pm 4.43$} 
& $\bm{69.29}$\tiny{$\pm 1.72$}\\

CIFAR100
& $30.54$\tiny{$\pm 0.46$} 
& $30.43$\tiny{$\pm 0.36$} 
& $32.10$\tiny{$\pm 0.58$} 
& $33.29$\tiny{$\pm 0.48$}  
& $43.27$\tiny{$\pm 1.06$} 
& $\bm{49.00}$\tiny{$\pm 1.88$} 
& $48.73$\tiny{$\pm 0.58$}\\

\bottomrule
\end{tabular} 
}
\end{table*}

We evaluate the performance of \ts{\algoname} in comparison to state-of-the-art (SOTA) neuron matching, knowledge distillation, and model fusion baselines across a range of image recognition tasks in a FL setting. 
The datasets that we use are (i) FashionMNIST \citep{xiao2017fmnist}, (ii) SVHN \citep{SVHN}, (iii) CIFAR10 \citep{kri2009cifar100} and (iv) CINIC10 \citep{darlow2018cinic}. Our code is available at \url{https://github.com/Divyansh03/FedFisher}.

\paragraph{Baselines.} We compare \ts{FedFisher} with 6 other one-shot baselines. \ts{FedAvg} is the de facto baseline in all our experiments. The other baselines that we compare with are (i)  \ts{PFNM} \citep{yurochkin2019bayesian, wang2020federated}, (ii) \ts{OTFusion} \citep{singh2020model}, (iii) \ts{DENSE} \citep{zhang2022dense}, (iv) \ts{FisherMerge} \citep{matena2022merging} and (v) \ts{RegMean} \citep{jin2022dataless}. \ts{PFNM} and \ts{OTFusion} are popular neuron matching methods that first permute the weights of local models and then average them, rather than directly averaging the weights. \ts{DENSE} is a SOTA method for one-shot FL based on data-free knowledge distillation that uses GANs to artificially generate data for distillation at the server. \ts{FisherMerge} and \ts{RegMean} are SOTA model fusion baselines. We note that, similar to \ts{\algonamed}, \ts{FisherMerge} also proposes using the diagonal Fisher when merging models. However, our application of the diagonal Fisher differs in how we compute the Fisher averaged model (\Cref{lemma:lemma_gd_proj}) and additional regularization (\Cref{eq:fisher_avg_sol}). We defer a more detailed discussion on the difference between the algorithms to \Cref{sec:appendix_related_work}. We avoid comparing with baselines that need auxiliary data or maintain an ensemble of local models \citep{guha2019one,diaotowards} at the server to ensure fairness of comparison. We also avoid comparing with algorithms that are inherently multi-round in nature such as \ts{FedProx} \citep{sahu2019federated}, \ts{SCAFFOLD} \citep{karimireddy2019scaffold} and adapative variants of \ts{FedAvg} such as \ts{FedAdam} \citep{reddi2020adaptive} since their performance would be similar to \ts{FedAvg} for one round.

\paragraph{Models and Other Details.} For FashionMNIST we use the LeNet architecture \citep{lecun1998gradient}; for other datasets we use a CNN model proposed in \cite{wang2020federated}. To simulate data heterogeneity among the client datasets, we split our original image dataset into $M$ partitions using a Dirichlet sampling procedure with parameter $\alpha$ \citep{hsu2019noniid,reddi2020adaptive}, with a lower value of $\alpha$ implying a more heterogeneous split. The local optimization procedure is the same across all algorithms. In particular, clients perform $E = 30$ epochs of local training using the SGD optimizer with local learning rate $\eta = 0.01$, batch size $64$ and momentum factor $0.9$. To compute the Fisher diagonal and Fisher K-FAC we use the \ts{nngeometry} package \citep{george2021nngeometry}. Further details, including how we tuned hyperparameters, can be found in \Cref{sec:appendix_addnl_expts}.

\subsection{Results}

\paragraph{\ts{\algoname} outperforms baselines across varying heterogeneity.} As a first step, we seek to understand the impact of data heterogeneity on the performance of \ts{\algoname} and other one-shot baselines. To do so, we fix the number of clients $M=5$ and vary $\alpha$  in the range $\{0.05, 0.1, 0.2\}$ which can be considered moderate to high data heterogeneity. \Cref{table:heterogeneity} summarizes the results obtained by the algorithms across the various datasets. We see that \ts{\algoname} variants, especially \ts{\algonamek}, consistently outperform baselines and give almost $10-20\%$ improvement over vanilla averaging in most cases. This highlights the effectiveness of \ts{\algoname} as a one-shot algorithm that can tackle data heterogeneity in FL settings while being computation and communication efficient. 

\paragraph{\ts{\algoname} outperforms baselines across varying number of clients.} In \Cref{sec:appendix_addnl_expts}, we provide experimental results in which we fix heterogeneity $\alpha = 0.3$ and vary the number of clients as $M = \{10,20,30\}$. Our results show that \ts{\algonamek} continues to outperform
baselines across different $M$ with up to $10\%$ improvement in some cases like CIFAR10.

\paragraph{Extension to Few-Shot Settings.} A natural question to consider is if we can gain additional utility from \ts{\algoname} by extending it to the \textit{few shot} setting, i.e., using a few more rounds of local training and aggregation. \Cref{fig:multi_round} presents the results of performing $5$ rounds of aggregation with different algorithms with $\alpha = 0.1$ and $M = 5$. We omit comparison with \ts{OTFusion} here because we did not find the performance to be competitive with \ts{FedAvg} after the second round. We also omit comparison with \ts{PFNM} as it modifies the global model architecture, leading to increased client computation and communication in every round. We see that while the use of multiple rounds improves the performance of all algorithms, \ts{\algoname} variants continue to show the greatest improvement, especially for CINIC10 which can be considered the hardest dataset in our experiments. This highlights the additional utility offered by \ts{\algoname} in few-shot settings.

\paragraph{Using Pre-trained Models.} As motivated by recent literature in FL \citep{nguyen2022begin, chen2022importance, tan2022federated}, in many cases the server might have access to a model pre-trained on a large public dataset. We consider a setting where the server has a ResNet-18 \citep{he2016deep} pre-trained on Tiny-ImageNet \citep{le2015tiny} and wants to fine-tune this model. The fine-tuning datasets that we consider are CIFAR10, CINIC10, GTSRB \cite{Stallkamp2011TheGT} and CIFAR100. In all cases we split the fine-tuning dataset across $M = 5$ clients with $\alpha = 0.1$ heterogeneity. We focus on full fine-tuning, i.e.,  clients update all weights in the model, for $E=30$ local epochs for CIFAR100 and $E=10$ local epochs for the rest. We also use a smaller step size of $\eta = 0.001$ in this setting. We omit comparison with \ts{PFNM} since it does not support ResNet-like architectures. \Cref{table:pretrained} summarizes the results of one-shot aggregation in this setting. We see that algorithms that use Fisher information, including \ts{FisherMerge}, improve on \ts{FedAvg} and other baselines by almost $20\%$, with our methods achieving the highest accuracy. We attribute this large improvement to the reduced distance between weights of the local models when starting from a pre-trained model, which in turn reduces the approximation error in Fisher averaging as discussed in \Cref{sec:theoretical_analysis}. 


\section{Conclusion}
In this work, we propose \ts{\algoname}, a novel algorithm for one-shot FL motivated by a Bayesian perspective of FL. We theoretically analyze \ts{\algoname} for two-layer over-parameterized neural networks and propose its practical versions \ts{\algonamed} and \ts{\algonamek} that outperform current state-of-the-art one-shot methods while being computation and communication efficient. As future work, we would like to extend the analysis of \ts{\algoname} for deeper neural networks and investigate the use of differential privacy to improve the privacy guarantees of the practical versions of \ts{\algoname}. 

\section*{Acknowledgments}
This work was supported in part by NSF grants CCF 2045694, CNS-2112471, CPS-2111751, SHF-2107024, ONR N00014-23-1-2149 and CMU Dean's and Barakat fellowships.

\bibliography{bibliography}

\begin{thebibliography}{72}
\providecommand{\natexlab}[1]{#1}
\providecommand{\url}[1]{\texttt{#1}}
\expandafter\ifx\csname urlstyle\endcsname\relax
  \providecommand{\doi}[1]{doi: #1}\else
  \providecommand{\doi}{doi: \begingroup \urlstyle{rm}\Url}\fi

\bibitem[Ainsworth et~al.(2023)Ainsworth, Hayase, and Srinivasa]{ainsworth2022git}
Samuel~K. Ainsworth, Jonathan Hayase, and Siddhartha~S. Srinivasa.
\newblock Git re-basin: Merging models modulo permutation symmetries.
\newblock In \emph{The Eleventh International Conference on Learning Representations}, 2023.

\bibitem[Akash et~al.(2022)Akash, Li, and Trillos]{akash2022wasserstein}
Aditya~Kumar Akash, Sixu Li, and Nicol{\'a}s~Garc{\'\i}a Trillos.
\newblock Wasserstein barycenter-based model fusion and linear mode connectivity of neural networks.
\newblock \emph{arXiv preprint arXiv:2210.06671}, 2022.

\bibitem[Al{-}Shedivat et~al.(2021)Al{-}Shedivat, Gillenwater, Xing, and Rostamizadeh]{al2020federated}
Maruan Al{-}Shedivat, Jennifer Gillenwater, Eric~P. Xing, and Afshin Rostamizadeh.
\newblock Federated learning via posterior averaging: {A} new perspective and practical algorithms.
\newblock In \emph{9th International Conference on Learning Representations}, 2021.

\bibitem[Allen-Zhu et~al.(2019{\natexlab{a}})Allen-Zhu, Li, and Liang]{allen2019learning}
Zeyuan Allen-Zhu, Yuanzhi Li, and Yingyu Liang.
\newblock Learning and generalization in overparameterized neural networks, going beyond two layers.
\newblock \emph{Advances in Neural Information Processing Systems}, 32, 2019{\natexlab{a}}.

\bibitem[Allen-Zhu et~al.(2019{\natexlab{b}})Allen-Zhu, Li, and Song]{allen2019convergence}
Zeyuan Allen-Zhu, Yuanzhi Li, and Zhao Song.
\newblock A convergence theory for deep learning via over-parameterization.
\newblock In \emph{International Conference on Machine Learning}, pages 242--252. PMLR, 2019{\natexlab{b}}.

\bibitem[Armacki et~al.(2022)Armacki, Bajovic, Jakovetic, and Kar]{armacki2022one}
Aleksandar Armacki, Dragana Bajovic, Dusan Jakovetic, and Soummya Kar.
\newblock One-shot federated learning for model clustering and learning in heterogeneous environments.
\newblock \emph{arXiv preprint arXiv:2209.10866}, 2022.

\bibitem[Arora et~al.(2019)Arora, Du, Hu, Li, and Wang]{arora2019fine}
Sanjeev Arora, Simon Du, Wei Hu, Zhiyuan Li, and Ruosong Wang.
\newblock Fine-grained analysis of optimization and generalization for overparameterized two-layer neural networks.
\newblock In \emph{International Conference on Machine Learning}, pages 322--332. PMLR, 2019.

\bibitem[Bonawitz et~al.(2016)Bonawitz, Ivanov, Kreuter, Marcedone, McMahan, Patel, Ramage, Segal, and Seth]{bonawitz2016practical}
Keith Bonawitz, Vladimir Ivanov, Ben Kreuter, Antonio Marcedone, H.~Brendan McMahan, Sarvar Patel, Daniel Ramage, Aaron Segal, and Karn Seth.
\newblock Practical secure aggregation for federated learning on user-held data.
\newblock In \emph{NeurIPS Workshop on Private Multi-Party Machine Learning}, 2016.

\bibitem[Chen et~al.(2022)Chen, Tu, Li, Shen, and Chao]{chen2022importance}
Hong-You Chen, Cheng-Hao Tu, Ziwei Li, Han~Wei Shen, and Wei-Lun Chao.
\newblock On the importance and applicability of pre-training for federated learning.
\newblock In \emph{The Eleventh International Conference on Learning Representations}, 2022.

\bibitem[Choshen et~al.(2022)Choshen, Venezian, Slonim, and Katz]{choshen2022fusing}
Leshem Choshen, Elad Venezian, Noam Slonim, and Yoav Katz.
\newblock Fusing finetuned models for better pretraining.
\newblock \emph{arXiv preprint arXiv:2204.03044}, 2022.

\bibitem[Darlow et~al.(2018)Darlow, Crowley, Antoniou, and Storkey]{darlow2018cinic}
Luke~N Darlow, Elliot~J Crowley, Antreas Antoniou, and Amos~J Storkey.
\newblock {CINIC-10} is not {Imagenet} or {CIFAR-10}.
\newblock \emph{arXiv preprint arXiv:1810.03505}, 2018.

\bibitem[Deng et~al.(2020)Deng, Kamani, and Mahdavi]{deng2020adaptive}
Yuyang Deng, Mohammad~Mahdi Kamani, and Mehrdad Mahdavi.
\newblock Adaptive personalized federated learning, 2020.

\bibitem[Dennis et~al.(2021)Dennis, Li, and Smith]{dennis2021heterogeneity}
Don~Kurian Dennis, Tian Li, and Virginia Smith.
\newblock Heterogeneity for the win: One-shot federated clustering.
\newblock In \emph{International Conference on Machine Learning}, pages 2611--2620. PMLR, 2021.

\bibitem[Diao et~al.(2023)Diao, Li, and He]{diaotowards}
Yiqun Diao, Qinbin Li, and Bingsheng He.
\newblock Towards addressing label skews in one-shot federated learning.
\newblock In \emph{The Eleventh International Conference on Learning Representations}, 2023.

\bibitem[Du et~al.(2019)Du, Zhai, P{\'{o}}czos, and Singh]{du2018gradient}
Simon~S. Du, Xiyu Zhai, Barnab{\'{a}}s P{\'{o}}czos, and Aarti Singh.
\newblock Gradient descent provably optimizes over-parameterized neural networks.
\newblock In \emph{7th International Conference on Learning Representations}, 2019.

\bibitem[Entezari et~al.(2022)Entezari, Sedghi, Saukh, and Neyshabur]{entezari2021role}
Rahim Entezari, Hanie Sedghi, Olga Saukh, and Behnam Neyshabur.
\newblock The role of permutation invariance in linear mode connectivity of neural networks.
\newblock In \emph{The Tenth International Conference on Learning Representations}, 2022.

\bibitem[Garin et~al.(2022)Garin, Evgeniou, and Vayatis]{garin2022weighting}
Marie Garin, Theodoros Evgeniou, and Nicolas Vayatis.
\newblock Weighting schemes for one-shot federated learning.
\newblock 2022.

\bibitem[George(2021)]{george2021nngeometry}
Thomas George.
\newblock {NNGeometry: Easy and Fast Fisher Information Matrices and Neural Tangent Kernels in PyTorch}, February 2021.
\newblock URL \url{https://doi.org/10.5281/zenodo.4532597}.

\bibitem[Gong et~al.(2021)Gong, Sharma, Karanam, Wu, Chen, Doermann, and Innanje]{gong2021ensemble}
Xuan Gong, Abhishek Sharma, Srikrishna Karanam, Ziyan Wu, Terrence Chen, David Doermann, and Arun Innanje.
\newblock Ensemble attention distillation for privacy-preserving federated learning.
\newblock In \emph{Proceedings of the IEEE/CVF International Conference on Computer Vision}, pages 15076--15086, 2021.

\bibitem[Grosse and Martens(2016)]{grosse2016kronecker}
Roger Grosse and James Martens.
\newblock A kronecker-factored approximate fisher matrix for convolution layers.
\newblock In \emph{International Conference on Machine Learning}, pages 573--582. PMLR, 2016.

\bibitem[Guha et~al.(2019)Guha, Talwalkar, and Smith]{guha2019one}
Neel Guha, Ameet Talwalkar, and Virginia Smith.
\newblock One-shot federated learning.
\newblock \emph{arXiv preprint arXiv:1902.11175}, 2019.

\bibitem[Guo et~al.(2023)Guo, Greengard, Wang, Gelman, Kim, and Xing]{guo2023federated}
Han Guo, Philip Greengard, Hongyi Wang, Andrew Gelman, Yoon Kim, and Eric~P. Xing.
\newblock Federated learning as variational inference: {A} scalable expectation propagation approach.
\newblock In \emph{The Eleventh International Conference on Learning Representations}, 2023.

\bibitem[He et~al.(2016)He, Zhang, Ren, and Sun]{he2016deep}
Kaiming He, Xiangyu Zhang, Shaoqing Ren, and Jian Sun.
\newblock Deep residual learning for image recognition.
\newblock In \emph{Proceedings of the IEEE conference on computer vision and pattern recognition}, pages 770--778, 2016.

\bibitem[Heinbaugh et~al.(2023)Heinbaugh, Luz-Ricca, and Shao]{heinbaughdata}
Clare~Elizabeth Heinbaugh, Emilio Luz-Ricca, and Huajie Shao.
\newblock Data-free one-shot federated learning under very high statistical heterogeneity.
\newblock In \emph{The Eleventh International Conference on Learning Representations}, 2023.

\bibitem[Hitaj et~al.(2017)Hitaj, Ateniese, and Perez-Cruz]{hitaj2017deep}
Briland Hitaj, Giuseppe Ateniese, and Fernando Perez-Cruz.
\newblock Deep models under the gan: information leakage from collaborative deep learning.
\newblock In \emph{Proceedings of the 2017 ACM SIGSAC conference on computer and communications security}, pages 603--618, 2017.

\bibitem[Hsu et~al.(2019)Hsu, Qi, and Brown]{hsu2019noniid}
Tzu-Ming~Harry Hsu, Hang Qi, and Matthew Brown.
\newblock Measuring the effects of non-identical data distribution for federated visual classification.
\newblock In \emph{International Workshop on Federated Learning for User Privacy and Data Confidentiality in Conjunction with NeurIPS 2019 (FL-NeurIPS'19)}, December 2019.

\bibitem[Huang et~al.(2021)Huang, Li, Song, and Yang]{huang2021fl}
Baihe Huang, Xiaoxiao Li, Zhao Song, and Xin Yang.
\newblock Fl-ntk: A neural tangent kernel-based framework for federated learning analysis.
\newblock In \emph{International Conference on Machine Learning}, pages 4423--4434. PMLR, 2021.

\bibitem[Ilharco et~al.(2023)Ilharco, Ribeiro, Wortsman, Schmidt, Hajishirzi, and Farhadi]{ilharco2022editing}
Gabriel Ilharco, Marco~T{\'{u}}lio Ribeiro, Mitchell Wortsman, Ludwig Schmidt, Hannaneh Hajishirzi, and Ali Farhadi.
\newblock Editing models with task arithmetic.
\newblock In \emph{The Eleventh International Conference on Learning Representations}, 2023.

\bibitem[Jin et~al.(2023)Jin, Ren, Preotiuc{-}Pietro, and Cheng]{jin2022dataless}
Xisen Jin, Xiang Ren, Daniel Preotiuc{-}Pietro, and Pengxiang Cheng.
\newblock Dataless knowledge fusion by merging weights of language models.
\newblock In \emph{The Eleventh International Conference on Learning Representations}, 2023.

\bibitem[Jordan et~al.(2023)Jordan, Sedghi, Saukh, Entezari, and Neyshabur]{jordan2022repair}
Keller Jordan, Hanie Sedghi, Olga Saukh, Rahim Entezari, and Behnam Neyshabur.
\newblock {REPAIR:} renormalizing permuted activations for interpolation repair.
\newblock In \emph{The Eleventh International Conference on Learning Representations}, 2023.

\bibitem[Kadhe et~al.(2020)Kadhe, Rajaraman, Koyluoglu, and Ramchandran]{kadhe2020fastsecagg}
Swanand Kadhe, Nived Rajaraman, O~Ozan Koyluoglu, and Kannan Ramchandran.
\newblock Fastsecagg: Scalable secure aggregation for privacy-preserving federated learning.
\newblock \emph{arXiv preprint arXiv:2009.11248}, 2020.

\bibitem[Kairouz et~al.(2021)Kairouz, McMahan, Avent, Bellet, Bennis, Bhagoji, Bonawitz, Charles, Cormode, Cummings, et~al.]{kairouz2019advances}
Peter Kairouz, H~Brendan McMahan, Brendan Avent, Aur{\'e}lien Bellet, Mehdi Bennis, Arjun~Nitin Bhagoji, Kallista Bonawitz, Zachary Charles, Graham Cormode, Rachel Cummings, et~al.
\newblock Advances and open problems in federated learning.
\newblock \emph{Foundations and Trends{\textregistered} in Machine Learning}, 14\penalty0 (1--2):\penalty0 1--210, 2021.

\bibitem[Karimireddy et~al.(2020)Karimireddy, Kale, Mohri, Reddi, Stich, and Suresh]{karimireddy2019scaffold}
Sai~Praneeth Karimireddy, Satyen Kale, Mehryar Mohri, Sashank Reddi, Sebastian Stich, and Ananda~Theertha Suresh.
\newblock Scaffold: Stochastic controlled averaging for federated learning.
\newblock In \emph{International Conference on Machine Learning}, pages 5132--5143. PMLR, 2020.

\bibitem[Kirkpatrick et~al.(2017)Kirkpatrick, Pascanu, Rabinowitz, Veness, Desjardins, Rusu, Milan, Quan, Ramalho, Grabska-Barwinska, et~al.]{kirkpatrick2017overcoming}
James Kirkpatrick, Razvan Pascanu, Neil Rabinowitz, Joel Veness, Guillaume Desjardins, Andrei~A Rusu, Kieran Milan, John Quan, Tiago Ramalho, Agnieszka Grabska-Barwinska, et~al.
\newblock Overcoming catastrophic forgetting in neural networks.
\newblock \emph{Proceedings of the national academy of sciences}, 114\penalty0 (13):\penalty0 3521--3526, 2017.

\bibitem[Krizhevsky et~al.(2009)Krizhevsky, Nair, and Hinton]{kri2009cifar100}
Alex Krizhevsky, Vinod Nair, and Geoffrey Hinton.
\newblock Cifar-100 (canadian institute for advanced research).
\newblock \url{http://www.cs.toronto.edu/~kriz/cifar.html}, 2009.

\bibitem[Kurach et~al.(2019)Kurach, Lu{\v{c}}i{\'c}, Zhai, Michalski, and Gelly]{kurach2019large}
Karol Kurach, Mario Lu{\v{c}}i{\'c}, Xiaohua Zhai, Marcin Michalski, and Sylvain Gelly.
\newblock A large-scale study on regularization and normalization in gans.
\newblock In \emph{International Conference on Machine Learning}, pages 3581--3590. PMLR, 2019.

\bibitem[Le and Yang(2015)]{le2015tiny}
Ya~Le and Xuan Yang.
\newblock Tiny imagenet visual recognition challenge.
\newblock \emph{CS 231N}, 7\penalty0 (7):\penalty0 3, 2015.

\bibitem[LeCun(1998)]{lecun1998mnist}
Yann LeCun.
\newblock The {MNIST} database of handwritten digits.
\newblock \emph{http://yann. lecun. com/exdb/mnist/}, 1998.

\bibitem[LeCun et~al.(1989)LeCun, Denker, and Solla]{lecun1989optimal}
Yann LeCun, John Denker, and Sara Solla.
\newblock Optimal brain damage.
\newblock \emph{Advances in Neural Information Processing Systems}, 2, 1989.

\bibitem[LeCun et~al.(1998)LeCun, Bottou, Bengio, and Haffner]{lecun1998gradient}
Yann LeCun, L{\'e}on Bottou, Yoshua Bengio, and Patrick Haffner.
\newblock Gradient-based learning applied to document recognition.
\newblock \emph{Proceedings of the IEEE}, 86\penalty0 (11):\penalty0 2278--2324, 1998.

\bibitem[Li et~al.(2021)Li, He, and Song]{li2020practical}
Qinbin Li, Bingsheng He, and Dawn Song.
\newblock Practical one-shot federated learning for cross-silo setting.
\newblock In \emph{Proceedings of the Thirtieth International Joint Conference on Artificial Intelligence (IJCAI-21)}, May 2021.

\bibitem[Li et~al.(2020{\natexlab{a}})Li, Sahu, Talwalkar, and Smith]{li2020federated}
Tian Li, Anit~Kumar Sahu, Ameet Talwalkar, and Virginia Smith.
\newblock Federated learning: Challenges, methods, and future directions.
\newblock \emph{IEEE Signal Processing Magazine}, 37\penalty0 (3):\penalty0 50--60, 2020{\natexlab{a}}.

\bibitem[Li et~al.(2020{\natexlab{b}})Li, Sahu, Zaheer, Sanjabi, Talwalkar, and Smith]{sahu2019federated}
Tian Li, Anit~Kumar Sahu, Manzil Zaheer, Maziar Sanjabi, Ameet Talwalkar, and Virginia Smith.
\newblock Federated optimization for heterogeneous networks.
\newblock In \emph{Proceedings of the 3rd MLSys Conference}, January 2020{\natexlab{b}}.

\bibitem[Lin et~al.(2020)Lin, Kong, Stich, and Jaggi]{lin2020ensemble}
Tao Lin, Lingjing Kong, Sebastian~U Stich, and Martin Jaggi.
\newblock Ensemble distillation for robust model fusion in federated learning.
\newblock \emph{Advances in Neural Information Processing Systems}, 33:\penalty0 2351--2363, 2020.

\bibitem[Liu et~al.(2022)Liu, Lou, Wang, Xi, Shen, and Yan]{liu2022deep}
Chang Liu, Chenfei Lou, Runzhong Wang, Alan~Yuhan Xi, Li~Shen, and Junchi Yan.
\newblock Deep neural network fusion via graph matching with applications to model ensemble and federated learning.
\newblock In \emph{International Conference on Machine Learning}, pages 13857--13869. PMLR, 2022.

\bibitem[Martens(2020)]{martens2020new}
James Martens.
\newblock New insights and perspectives on the natural gradient method.
\newblock \emph{The Journal of Machine Learning Research}, 21\penalty0 (1):\penalty0 5776--5851, 2020.

\bibitem[Martens and Grosse(2015)]{martens2015optimizing}
James Martens and Roger Grosse.
\newblock Optimizing neural networks with kronecker-factored approximate curvature.
\newblock In \emph{International Conference on Machine Learning}, pages 2408--2417. PMLR, 2015.

\bibitem[Matena and Raffel(2022)]{matena2022merging}
Michael~S Matena and Colin~A Raffel.
\newblock Merging models with fisher-weighted averaging.
\newblock \emph{Advances in Neural Information Processing Systems}, 35:\penalty0 17703--17716, 2022.

\bibitem[McMahan et~al.(2017)McMahan, Moore, Ramage, Hampson, and y~Arcas]{mcmahan2017communication}
H.~Brendan McMahan, Eider Moore, Daniel Ramage, Seth Hampson, and Blaise~Ag\o{u}ra y~Arcas.
\newblock {Communication-Efficient Learning of Deep Networks from Decentralized Data}.
\newblock \emph{International Conference on Artificial Intelligenece and Statistics (AISTATS)}, April 2017.
\newblock URL \url{https://arxiv.org/abs/1602.05629}.

\bibitem[Mohri et~al.(2018)Mohri, Rostamizadeh, and Talwalkar]{mohri2018foundations}
Mehryar Mohri, Afshin Rostamizadeh, and Ameet Talwalkar.
\newblock \emph{Foundations of machine learning}.
\newblock MIT press, 2018.

\bibitem[Netzer et~al.(2011)Netzer, Wang, Coates, Bissacco, Wu, and Ng]{SVHN}
Yuval Netzer, Tao Wang, Adam Coates, Alessandro Bissacco, Bo~Wu, and Andrew~Y. Ng.
\newblock Reading digits in natural images with unsupervised feature learning.
\newblock In \emph{NeurIPS Workshop on Deep Learning and Unsupervised Feature Learning}, 2011.

\bibitem[Nguyen et~al.(2022)Nguyen, Malik, Sanjabi, and Rabbat]{nguyen2022begin}
John Nguyen, Kshitiz Malik, Maziar Sanjabi, and Michael Rabbat.
\newblock Where to begin? exploring the impact of pre-training and initialization in federated learning.
\newblock \emph{arXiv preprint arXiv:2206.15387}, 2022.

\bibitem[Reddi et~al.(2021)Reddi, Charles, Zaheer, Garrett, Rush, Kone{\v{c}}n{\`y}, Kumar, and McMahan]{reddi2020adaptive}
Sashank Reddi, Zachary Charles, Manzil Zaheer, Zachary Garrett, Keith Rush, Jakub Kone{\v{c}}n{\`y}, Sanjiv Kumar, and H~Brendan McMahan.
\newblock Adaptive federated optimization.
\newblock In \emph{International Conference on Learning Representations (ICLR)}, 2021.

\bibitem[Shin et~al.(2020)Shin, Hwang, Kim, Park, Bennis, and Kim]{shin2020xor}
MyungJae Shin, Chihoon Hwang, Joongheon Kim, Jihong Park, Mehdi Bennis, and Seong-Lyun Kim.
\newblock Xor mixup: Privacy-preserving data augmentation for one-shot federated learning.
\newblock \emph{arXiv preprint arXiv:2006.05148}, 2020.

\bibitem[Singh and Alistarh(2020)]{singh2020woodfisher}
Sidak~Pal Singh and Dan Alistarh.
\newblock Woodfisher: Efficient second-order approximation for neural network compression.
\newblock \emph{Advances in Neural Information Processing Systems}, 33:\penalty0 18098--18109, 2020.

\bibitem[Singh and Jaggi(2020)]{singh2020model}
Sidak~Pal Singh and Martin Jaggi.
\newblock Model fusion via optimal transport.
\newblock \emph{Advances in Neural Information Processing Systems}, 33:\penalty0 22045--22055, 2020.

\bibitem[Song et~al.()Song, Khanduri, Zhang, Yi, and Hong]{songfedavg}
Bingqing Song, Prashant Khanduri, Xinwei Zhang, Jinfeng Yi, and Mingyi Hong.
\newblock Fedavg converges to zero training loss linearly: The power of overparameterized multi-layer neural networks.

\bibitem[Stallkamp et~al.(2011)Stallkamp, Schlipsing, Salmen, and Igel]{Stallkamp2011TheGT}
Johannes Stallkamp, Marc Schlipsing, Jan Salmen, and C.~Igel.
\newblock The german traffic sign recognition benchmark: A multi-class classification competition.
\newblock \emph{The 2011 International Joint Conference on Neural Networks}, pages 1453--1460, 2011.
\newblock URL \url{https://api.semanticscholar.org/CorpusID:15926837}.

\bibitem[Tan et~al.(2022)Tan, Long, Ma, Liu, Zhou, and Jiang]{tan2022federated}
Yue Tan, Guodong Long, Jie Ma, Lu~Liu, Tianyi Zhou, and Jing Jiang.
\newblock Federated learning from pre-trained models: A contrastive learning approach.
\newblock \emph{Advances in Neural Information Processing Systems}, 35:\penalty0 19332--19344, 2022.

\bibitem[Wang et~al.(2020{\natexlab{a}})Wang, Yurochkin, Sun, Papailiopoulos, and Khazaeni]{wang2020federated}
Hongyi Wang, Mikhail Yurochkin, Yuekai Sun, Dimitris~S. Papailiopoulos, and Yasaman Khazaeni.
\newblock Federated learning with matched averaging.
\newblock In \emph{8th International Conference on Learning Representations}, 2020{\natexlab{a}}.

\bibitem[Wang et~al.(2020{\natexlab{b}})Wang, Liu, Liang, Joshi, and Poor]{wang2020tackling}
Jianyu Wang, Qinghua Liu, Hao Liang, Gauri Joshi, and H~Vincent Poor.
\newblock Tackling the objective inconsistency problem in heterogeneous federated optimization.
\newblock \emph{Advances in Neural Information Processing Systems}, 33:\penalty0 7611--7623, 2020{\natexlab{b}}.

\bibitem[Xiao et~al.(2017)Xiao, Rasul, and Vollgraf]{xiao2017fmnist}
Han Xiao, Kashif Rasul, and Roland Vollgraf.
\newblock Fashion-mnist: a novel image dataset for benchmarking machine learning algorithms.
\newblock \emph{https://arxiv.org/abs/1708.07747}, aug 2017.

\bibitem[Yadav et~al.(2023)Yadav, Tam, Choshen, Raffel, and Bansal]{yadav2023resolving}
Prateek Yadav, Derek Tam, Leshem Choshen, Colin Raffel, and Mohit Bansal.
\newblock Resolving interference when merging models.
\newblock \emph{arXiv preprint arXiv:2306.01708}, 2023.

\bibitem[Yang et~al.(2019)Yang, Liu, Chen, and Tong]{yang2019federated}
Qiang Yang, Yang Liu, Tianjian Chen, and Yongxin Tong.
\newblock Federated machine learning: Concept and applications.
\newblock \emph{ACM Transactions on Intelligent Systems and Technology (TIST)}, 10\penalty0 (2):\penalty0 12, 2019.

\bibitem[Yu et~al.(2022)Yu, Wei, Karimireddy, Ma, and Jordan]{yu2022tct}
Yaodong Yu, Alexander Wei, Sai~Praneeth Karimireddy, Yi~Ma, and Michael Jordan.
\newblock Tct: Convexifying federated learning using bootstrapped neural tangent kernels.
\newblock \emph{Advances in Neural Information Processing Systems}, 35:\penalty0 30882--30897, 2022.

\bibitem[Yue et~al.(2022)Yue, Jin, Pilgrim, Wong, Baron, and Dai]{yue2022neural}
Kai Yue, Richeng Jin, Ryan Pilgrim, Chau-Wai Wong, Dror Baron, and Huaiyu Dai.
\newblock Neural tangent kernel empowered federated learning.
\newblock In \emph{International Conference on Machine Learning}, pages 25783--25803. PMLR, 2022.

\bibitem[Yurochkin et~al.(2019)Yurochkin, Agarwal, Ghosh, Greenewald, Hoang, and Khazaeni]{yurochkin2019bayesian}
Mikhail Yurochkin, Mayank Agarwal, Soumya Ghosh, Kristjan Greenewald, Nghia Hoang, and Yasaman Khazaeni.
\newblock Bayesian nonparametric federated learning of neural networks.
\newblock In \emph{International Conference on Machine Learning}, pages 7252--7261. PMLR, 2019.

\bibitem[Zhang et~al.(2022)Zhang, Chen, Li, Lyu, Wu, Ding, Shen, and Wu]{zhang2022dense}
Jie Zhang, Chen Chen, Bo~Li, Lingjuan Lyu, Shuang Wu, Shouhong Ding, Chunhua Shen, and Chao Wu.
\newblock Dense: Data-free one-shot federated learning.
\newblock \emph{Advances in Neural Information Processing Systems}, 35:\penalty0 21414--21428, 2022.

\bibitem[Zhang et~al.(2020)Zhang, Jia, Pei, Wang, Li, and Song]{zhang2020secret}
Yuheng Zhang, Ruoxi Jia, Hengzhi Pei, Wenxiao Wang, Bo~Li, and Dawn Song.
\newblock The secret revealer: Generative model-inversion attacks against deep neural networks.
\newblock In \emph{Proceedings of the IEEE/CVF conference on computer vision and pattern recognition}, pages 253--261, 2020.

\bibitem[Zhou et~al.(2020)Zhou, Pu, Ma, Li, and Wu]{zhou2020distilled}
Yanlin Zhou, George Pu, Xiyao Ma, Xiaolin Li, and Dapeng Wu.
\newblock Distilled one-shot federated learning.
\newblock \emph{arXiv preprint arXiv:2009.07999}, 2020.

\bibitem[Zhu et~al.(2021)Zhu, Hong, and Zhou]{zhu2021data}
Zhuangdi Zhu, Junyuan Hong, and Jiayu Zhou.
\newblock Data-free knowledge distillation for heterogeneous federated learning.
\newblock In \emph{International Conference on Machine Learning}, pages 12878--12889. PMLR, 2021.

\bibitem[Zou and Gu(2019)]{zou2019improved}
Difan Zou and Quanquan Gu.
\newblock An improved analysis of training over-parameterized deep neural networks.
\newblock \emph{Advances in Neural Information Processing Systems}, 32, 2019.

\end{thebibliography}

\section*{Checklist}

 \begin{enumerate}

 \item For all models and algorithms presented, check if you include:
 \begin{enumerate}
   \item A clear description of the mathematical setting, assumptions, algorithm, and/or model. [Yes/No/Not Applicable]
   
   \textbf{Yes. Please see \Cref{sec:propose_algo} and \Cref{sec:theoretical_analysis}.}
   \item An analysis of the properties and complexity (time, space, sample size) of any algorithm. [Yes/No/Not Applicable]
   
   \textbf{Yes. Please see \Cref{sec:theoretical_analysis} and \Cref{sec:practical_algo}.}
   \item (Optional) Anonymized source code, with specification of all dependencies, including external libraries. [Yes/No/Not Applicable]

   \textbf{Yes. Please see supplemental.}
 \end{enumerate}

 \item For any theoretical claim, check if you include:
 \begin{enumerate}
   \item Statements of the full set of assumptions of all theoretical results. [Yes/No/Not Applicable]

   \textbf{Yes. Please see \Cref{sec:theoretical_analysis}.}

   \item Complete proofs of all theoretical results.
   [Yes/No/Not Applicable]

   \textbf{Yes. Please see \Cref{sec:appendix_proofs}.}

   \item Clear explanations of any assumptions. [Yes/No/Not Applicable]

   \textbf{Yes. Please see \Cref{sec:theoretical_analysis}.}
   
 \end{enumerate}

 \item For all figures and tables that present empirical results, check if you include:
 \begin{enumerate}
   \item The code, data, and instructions needed to reproduce the main experimental results (either in the supplemental material or as a URL). [Yes/No/Not Applicable]

   \textbf{Yes. Please see \Cref{sec:experiments} and \Cref{sec:appendix_addnl_expts} in supplemental.}

   \item All the training details (e.g., data splits, hyperparameters, how they were chosen). [Yes/No/Not Applicable]

   \textbf{Yes. Please see \Cref{sec:experiments} and \Cref{sec:appendix_addnl_expts} in supplemental.}

     \item A clear definition of the specific measure or statistics and error bars (e.g., with respect to the random seed after running experiments multiple times). 

     \textbf{Yes. Please see \Cref{sec:appendix_addnl_expts} in supplemental.}

     \item A description of the computing infrastructure used. (e.g., type of GPUs, internal cluster, or cloud provider). [Yes/No/Not Applicable]

      \textbf{Yes. Please see \Cref{sec:appendix_addnl_expts}.}
     
 \end{enumerate}

 \item If you are using existing assets (e.g., code, data, models) or curating/releasing new assets, check if you include:
 \begin{enumerate}
   \item Citations of the creator If your work uses existing assets. [Yes/No/Not Applicable]

   \textbf{Yes.}
   
   \item The license information of the assets, if applicable. [Yes/No/Not Applicable]

   \textbf{Not Applicable.}

   \item New assets either in the supplemental material or as a URL, if applicable. [Yes/No/Not Applicable]

   \textbf{Not Applicable.}
   
   \item Information about consent from data providers/curators. [Yes/No/Not Applicable]

   \textbf{Not Applicable.}
   
   \item Discussion of sensible content if applicable, e.g., personally identifiable information or offensive content. [Yes/No/Not Applicable]

   \textbf{Not Applicable.}
   
 \end{enumerate}

 \item If you used crowdsourcing or conducted research with human subjects, check if you include:
 \begin{enumerate}
   \item The full text of instructions given to participants and screenshots. [Yes/No/Not Applicable]

   \textbf{Not Applicable.}
   
   \item Descriptions of potential participant risks, with links to Institutional Review Board (IRB) approvals if applicable. [Yes/No/Not Applicable]

   \textbf{Not Applicable.}
   \item The estimated hourly wage paid to participants and the total amount spent on participant compensation. [Yes/No/Not Applicable]

   \textbf{Not Applicable.}
 \end{enumerate}

 \end{enumerate}

\appendix
\setlength{\footskip}{20pt}
\pagestyle{plain}
\onecolumn
\aistatstitle{Appendix for ``FedFisher: Leveraging Fisher Information for One-Shot Federated Learning''}

\startcontents[sections]
\printcontents[sections]{l}{1}

\setcounter{table}{3}
\setcounter{figure}{2}
\setcounter{prop}{0}
\setcounter{lem}{0}
\setcounter{thm}{0}
\setcounter{defn}{1}
\setcounter{assum}{4}
\renewcommand\thesection{\Alph{section}}

\vfill

\newpage

\section{Additional Related Work}
\label{sec:appendix_related_work}
\paragraph{One-Shot FL.} We review here some additional work on one-shot FL apart from the Knowledge Distillation and Neuron Matching baselines discussed in our work. Initial works such as \cite{guha2019one} propose to just use the ensemble of client models at the server. However, this approach increases the storage and computation cost by a factor of $M$ where $M$ is the number of clients. The work of \cite{diaotowards} discusses how we can improve the prediction of this ensemble when the label distribution across client data is highly skewed. Another line of work proposes that clients send some distilled form of their data to the server \citep{zhou2020distilled,shin2020xor}. However, the privacy guarantees of such methods is unclear. The work of \cite{garin2022weighting} proposes techniques to optimize the weights of given to the local models when aggregating them at the server to improve one-shot performance. However, their analysis is limited to simple linear models and does not consider combing neural networks. We also note the existence of works that propose to perform clustering in a one-shot manner in FL setting \citep{armacki2022one,dennis2021heterogeneity}; these approaches our orthogonal to our problem of finding a global minimizer in a one-shot manner.

\paragraph{Convergence of overparameterized NNs in FL.}
The works of \cite{huang2021fl}, \cite{deng2020adaptive,songfedavg} study the convergence of \ts{FedAvg} for overparameterized neural networks. We note that these works are primarily concerned with convergence and do not propose any new algorithms as such compared to our work. We also note the existence of related works \citep{yu2022tct,yue2022neural} that proposes to use NTK style Jacobian features to speed up FL training; however these works usually require multiple training rounds.

\paragraph{Comparison with \citep{matena2022merging}.} We note that, similar to \ts{\algonamed}, \ts{FisherMerge} \citep{matena2022merging} also proposes using the diagonal Fisher when merging models. However, there are several key differences which we would like to mention. Firstly our problem is more well-defined in the sense that we have an explicit objective in Equation (1) that we are trying to solve with Fisher merging. This theoretical formulation is what leads us to propose the additional regularization in \ts{\algoname} in Equation (7) (and also \ts{\algonamed} subsequently) and guarantees for two-layer networks. \cite{matena2022merging} on the other hand does not provide any theoretical guarantees. Also note that the procedure to compute the Fisher averaged model is different (gradient descent in our work versus diagonal Fisher inversion in \cite{matena2022merging}). 
Finally, on a practical side, \cite{matena2022merging} do not consider the effect of data heterogeneity in their experiments and focus on fusing two models whereas we explicitly account for data heterogeneity in our model training and aggregate up to thirty local models in our experiments (see results in \Cref{fig:client_figure}).

\vfill

\newpage

\section{Proofs}
\label{sec:appendix_proofs}
\subsection{Proof of \Cref{prop:posterior_decomp}.}

\begin{prop} (Global Posterior Decomposition \citep{al2020federated})
Under the flat prior $\prob(\bW) \propto 1$, the global posterior decomposes into a product of local posteriors, i.e., $\prob(\bW|\mathcal{D}) \propto \prod_{i=1}^M \prob(\bW|\mathcal{D}_i)$. Furthermore, modes of the global posterior coincide with the optima of the FL objective in Equation (1), i.e, $\argmax_{\bW \in \mathbb{R}^d} \prob(\bW|\mathcal{D}) = \argmin_{\bW \in \mathbb{R}^d} L(\bW)$. 
\end{prop}

\textbf{Proof.}

We have,
\begin{align}
    \prob(\bW|\mathcal{D}) &\propto \prob(\mathcal{D}|\bW) && (\because \prob(\bW) \propto 1) \nonumber \\
    & = \prod_{i=1}^M \prob(\mathcal{D}_i|\bW) &&(\mathcal{D}_i \text{ are i.i.d generated}) \nonumber \\
    &\propto \prod_{i=1}^M \prob(\bW|\mathcal{D}_i).
\end{align}
Also,
\begin{align}
    \argmax_{\bW \in \mathbb{R}^d} \prob(\bW|\mathcal{D}) & = \argmax_{\bW \in \mathbb{R}^d} \log \prob(\mathcal{D}|\bW) \nonumber \\
    & = \argmax_{\bW \in \mathbb{R}^d} \sum_{i=1}^M\sum_{j=1}^n\left( \log q(\bx_{ij}) - \ell(f(\bW,\bx_{ij}),\by_{ij})\right) && (\Cref{assum:data_normalization}) \nonumber \\
    & = \argmax_{\bW \in \mathbb{R}^d} - \sum_{i=1}^M\sum_{j=1}^n \ell(f(\bW,\bx_{ij}),\by_{ij}) \nonumber \\
    & = \argmin_{\bW \in \mathbb{R}^d} L(\bW).
\end{align}
\qed

\subsection{Proof of \Cref{prop:posterior_solution}.}

\begin{prop}
Any $\bW$ which satisfies $\left(\sum_{i=1}^M \afi\right)\bW = \sum_{i=1}^M \afi \tmi$ is a minimizer of the objective $\sum_{i=1}^M (\bW - \tmi)^{\top}\afi(\bW-\tmi)$. 
\end{prop}

\textbf{Proof.}

Define $G(\bW) = \sum_{i=1}^M (\bW - \tmi)^{\top}\afi(\bW-\tmi)$. First note that 
\begin{align}
    \nabla^2 G(\bW) = 2\sum_{i=1}^M \afi \succcurlyeq 0 && (\text{$\afi$ is positive semi-definite and symmetric $\forall i \in [M]$).} \label{eq:f_convex_prof} 
\end{align}
This implies that $G(\bW)$ is a convex function. Therefore any $\bW$ which satisfies $\nabla G(\bW) = 0$ is a minimizer of $G(\bW)$. We have,
\begin{align}
    \nabla G(\bW) = 2\sum_{i=1}^M \afi(\bW - \bW_i).
\end{align}
Setting $\nabla G(\bW) = 0$ we get $\left(\sum_{i=1}^M \afi\right)\bW = \sum_{i=1}^M \afi \tmi$.
\qed

\subsection{Proof of \Cref{lemma:lemma_gd_proj}.}

\begin{lem}
Let $\bW^{(1)},\bW^{(2)},\dots$ be the iterates generated by running the following gradient descent (GD) procedure: $\bW^{(t+1)} = \bW^{(t)} - \eta_S \sum_{i=1}^M\left(\afi\bW^{(t)} - \afi\tmi\right)$ with $\bW^{(0)} = \sum_{i=1}^M \tmi/M$ and $\eta_S \leq 1/\lambda_{\max}$ where $\lambda_{\max}$ is the maximum eigenvalue of $\sum_{i=1}^M \afi$. Then, $\lim_{T \rightarrow \infty} \bW^{(T)} = \bW^*$.
\end{lem}

\textbf{Proof.}

Recall,

\begin{align}
  \label{eq:fisher_avg_sol_reperat}
    \omi = \argmin_{\bW \in \mathbb{R}^d}\left\{\tilde{L}(\bW) = \sum_{i=1}^M \norm{\bW - \tmi}^2 \text{ such that } \left(\sum_{i=1}^M \afi\right)\bW = \sum_{i=1}^M\afi \tmi\right\}. 
\end{align}

Let $\bF = \sum_{i=1}^M \afi$, $\bb = \sum_{i=1}^M \afi \tmi$ and $\bar{\bW} = \sum_{i=1}^M \tmi/M$. Also note that $\sum_{i=1}^M \norm{\bW - \tmi} = M \norm{\bW - \bar{\bW}}^2 + \sum_{i=1}^M \norm{\bar{\bW} - \tmi}^2$. Therefore, the expression for $\omi$ can be simplified as,
\begin{align}
    \omi = \argmin_{\bW \in \mathbb{R}^d}\left\{\tilde{L}(\bW) = \norm{\bW - \bar{\bW}}^2 \text{ such that } \bF\bW = \bb\right\}.
\end{align}
Since $\bF$ is symmetric and PSD (see \Cref{eq:f_convex_prof}), we have by the spectral decomposition of $\bF$,
\begin{align}
\bF 
= \bV\bSigma\bV^{\top} 
= \begin{bmatrix}
\bV_1 & \bV_2
\end{bmatrix}
\begin{bmatrix}
\bSigma_1 & \bm{0}\\
\bm{0} & \bm{0}
\end{bmatrix}
\begin{bmatrix}
\bV_1^{\top}\\
\bV_2^{\top}
\end{bmatrix}
& = \bV_1 \bSigma_1 \bV_1^{\top}.
\end{align}
Here $\bV \in \mathbb{R}^{(d \times d)}$ is an orthogonal matrix consisting of the eigenvectors of $\bF$, and $\bSigma$ is a diagonal matrix consisting of all the eigenvalues of $\bF$. $\bSigma_1$ is a diagonal matrix consisting of only the positive eigenvalues. $\bV_1$ consists of the eigenvectors corresponding to the positive eigenvalues while $\bV_2$ consists of the eigenvectors corresponding to the zero eigenvalues. Also note that we have $\bV_1^{\top}\bV_2 = \mathbf{0}$.

We first observe that for any $\bW_1^*$ and $\bW_2^*$ such that $\bF\bW_1^* = \bF\bW_2^* = \bb$ we have,
\begin{align}
    \bF\bW_1^* = \bF\bW_2^* & \iff  \bV_1 \bSigma_1 \bV_1^{\top}\bW_1^* = \bV_1 \bSigma_1 \bV_1^{\top}\bW_2^* \nonumber \\
   & \iff \left(\bV_1 \bSigma_1^{-1} \bV_1^{\top} \right)\bV_1 \bSigma_1 \bV_1^{\top}\bW_1^* = \left(\bV_1 \bSigma_1^{-1} \bV_1^{\top} \right)\bV_1 \bSigma_1 \bV_1^{\top}\bW_2^* \nonumber \\
   & \iff \bV_1\bV_1^{\top}\bW_1^* = \bV_1 \bV_1^{\top}\bW_2^*. \label{eq:lemma_1_proof_1}
\end{align}

Next we observe,
\begin{align}
    \bV_2\bV_2^{\top}\bW^* = \bV_2\bV_2^{\top}\bar{\bW}. \label{eq:lemma_1_proof_2}
\end{align}

The proof for this follows via simple contradiction argument as follows. Suppose $\bV_2\bV_2^{\top}\bW^* \neq \bV_2\bV_2^{\top}\bar{\bW}$. Let $\osmi$ be a vector such that $\bF\osmi = \bb$ and $\bV_2\bV_2^{\top}\osmi = \bV_2\bV_2^{\top}\bar{\bW}$. Then, 
\begin{align}
    \norm{\bar{\bW} - \omi}^2 &= \norm{\bV_2\bV_2^\top\left( \bar{\bW} - \omi\right) + \bV_1\bV_1^\top\left( \bar{\bW} - \omi\right)}^2 &&(\bV_2\bV_2^{\top} + \bV_1\bV_1^{\top} = \bI) \nonumber\\
    &= \norm{\bV_2\bV_2^\top\left( \bar{\bW} - \omi\right) + \bV_1\bV_1^\top\left( \bar{\bW} - \osmi\right)}^2 &&(\text{from \Cref{eq:lemma_1_proof_1}}) \nonumber\\
    & = \norm{\bV_2\bV_2^\top\left( \bar{\bW} - \omi\right)}^2 + \norm{\bV_1\bV_1^\top\left( \bar{\bW} - \osmi\right)}^2 &&(\bV_1^{\top}\bV_2 = \mathbf{0}) \nonumber\\
    & = \norm{\bV_2\bV_2^\top\left( \bar{\bW} - \omi\right)}^2 + \norm{\bar{\bW} - \osmi}^2 && (\bV_2\bV_2^{\top}(\bar{\bW} - \osmi) = 0 ) \nonumber\\
    & > \norm{\bar{\bW} - \osmi}^2 && ( \bV_2\bV_2^{\top}\bW^* \neq \bV_2\bV_2^{\top}\bar{\bW} ).
\end{align}
leading to a contradiction. 

According to the GD step in \Cref{lemma:lemma_gd_proj} we have,

\begin{align}
    \bW^{(t+1)} &= \bW^{(t)} - \eta_S(\bF\bW^{(t)} - \bb) \nonumber \\
    & = (\bI - \eta_S\bF)\bW^{(t)}+ \eta_S\bb .
\end{align}
Therefore,
\begin{align}
    \bW^{(T)} &= (\bI - \eta_S\bF)^{T}\bW^{(0)}+ \eta_S\sum_{t=0}^{T-1}(\bI - \eta_S\bF)^{t}\bb \nonumber \\
    & = (\bI - \eta_S\bF)^{T}\bW^{(0)}+ \eta_S\sum_{t=0}^{T-1}(\bI - \eta_S\bF)^{t}\bF\bW^* \nonumber \\
    & = \bV(\bI - \eta_S\bSigma)^{T}\bV^{\top}\bW^{(0)} + \eta_S\sum_{t=0}^{T-1}\left(\bV(\bI - \eta_S\bSigma)^{t}\bV^{\top}\right)\bV_1\bSigma_1\bV_1^{\top}\bW^* \nonumber \\
    & = (\bV_1 (\bI - \eta_S \bSigma_1)^T\bV_1 + \bV_2\bV_2^{\top})\bW^{(0)} + \eta_S \left(\bV_1 \sum_{t=0}^{T-1}(\bI - \eta_S \bSigma_1)^t\bV_1 + \bV_2\bV_2^{\top}\right) \bV_1\bSigma_1\bV_1^{\top}\bW^* \nonumber \\
    & = (\bV_1 (\bI - \eta_S \bSigma_1)^T\bV_1^{\top} + \bV_2\bV_2^{\top})\bW^{(0)} + \eta_S \left(\bV_1 \sum_{t=0}^{T-1}(\bI - \eta_S \bSigma_1)^t\bV_1^{\top} \right)\bV_1\bSigma_1\bV_1^{\top}\bW^*.
    \end{align}

In the limit $T \rightarrow \infty$ and with $\eta_S \leq 1/\lambda_{\max}(\bSigma_1)$, we have,
\begin{align}
    \lim_{T \rightarrow \infty} (\bI - \eta_S \bSigma_1)^T = \bm{0} \text{ and } \lim_{T \rightarrow \infty} \sum_{t=0}^{T-1}(\bI - \eta_S \bSigma_1)^t = \frac{1}{\eta_S}\bSigma_{1}^{-1}.
\end{align}
Thus,
\begin{align}
    \lim_{T \rightarrow \infty} \bW^{(T)} &= \bV_2 \bV_2^{\top}\bW^{(0)}+ \bV_1\bV_1^{\top}\bW^* \nonumber\\
    & = \bV_2 \bV_2^{\top}\bar{\bW}+ \bV_1\bV_1^{\top}\bW^* \nonumber \\
    & = \bV_2 \bV_2^{\top}\bW^*+ \bV_1\bV_1^{\top}\bW^* && (\text{from \Cref{eq:lemma_1_proof_2}}) \nonumber \\
    & = \bW^*.
\end{align}
This completes the proof.
\qed

\vfill

\newpage

\subsection{Proof of \Cref{theorem:fisher_avg_error}}

In this subsection, we provide the proof for \Cref{theorem:fisher_avg_error} in \Cref{sec:theoretical_analysis} of our work. To do so, we first introduce some additional notation and basic results.

\label{subsection:definitions}

\subsubsection{Additional Notation and Basic Results}

Recall the two-layer ReLU NN is modeled as follows,
\begin{align}
     f(\bW,\bx) = \frac{1}{\sqrt{m}}\sum_{r=1}^m a_r \bx^{\top}\bw_r \ind{\bx^{\top}\bw_r \geq 0}.
\end{align}

We can write the output of the neural network alternatively as,
\begin{align}
\label{eq:alternate_output_exp}
    f(\bW,\bx) &= \sum_{r=1}^m \left( \frac{1}{\sqrt{m}} a_r \ind{\bx^{\top}\bw_r \geq 0} \bx\right)^{\top}\bw_r \nonumber \\
    & = \phi(\bW,\bx)^{\top}\bW,
\end{align}

where
\begin{align}
    \phi(\bW,\bx) = \frac{1}{\sqrt{m}}
    \begin{bmatrix}
    a_1\bx \ind{\bx^{\top}\bw_1 \geq 0}\\ a_2\bx\ind{\bx^{\top}\bw_2 \geq 0}\\
    \vdots\\ 
    a_m\bx\ind{\bx^{\top}\bw_m \geq 0}]
    \end{bmatrix}
    \in \mathbb{R}^{mp}.
\end{align}

\paragraph{Initialization.}
Recall $\bW_0$ is the common initialization point for all the local models before they perform local optimization, i.e., $\bW_i^{(0)} = \bW_0$. Also recall $\bW^{(0)} = \bar{\bW} = \sum_{i=1}^M \tmi/M$ is the initialization point for the server optimization.

\begin{defn} (Matrices $\bA_{0}$, $\tilde{\bA}$, $\bH_{0}$, $\tilde{\bH}$) 
\label{defn:tilde_A}
Matrices $\bA_0$ and $\tilde{\bA}$ are defined as follows:
 \begin{align}
     \bA_{0} = \begin{bmatrix}
         \phi(\bW_0,\bx_{11})^{\top}\\
         \phi(\bW_0,\bx_{12})^{\top}\\
         \vdots\\
         \phi(\bW_0,\bx_{Mn})^{\top}
     \end{bmatrix} \in \mathbb{R}^{(N \times mp)}
 \end{align}
 and,
 \begin{align}
     \tilde{\bA} = \begin{bmatrix}
     \phi(\tilde{\bW}_1,\bx_{11})^{\top}\\
         \vdots\\
         \phi(\tilde{\bW}_M,\bx_{Mn})^{\top}
     \end{bmatrix} \in \mathbb{R}^{(N \times mp)}.
 \end{align}
 
Furthermore we define $\bH_{0} = \bA_{0}\bA_{0}^{\top}$, $\tilde{\bH} = \tilde{\bA}\tilde{\bA}^{\top} \in \mathbb{R}^{N \times N}$.

\end{defn}

\begin{defn}(Vector of true labels $\by$ and vector of predicted outputs $\tilde{\by}$)

The vector of true labels $\by$ is defined as follows,
\begin{align}
    \by = \begin{bmatrix}
        y_{11}\\
        y_{12}\\
        \vdots\\
        y_{Mn}
    \end{bmatrix} \in \mathbb{R}^N.
\end{align}

Given local models $\tilde{\bW}_1, \tilde{\bW}_2,\dots, \tilde{\bW}_M$, the vector of predicted outputs $\truey$ is defined as follows,
\begin{align}
    \truey = \begin{bmatrix}
        f(\tilde{\bW}_1,\bx_{11})\\
        f(\tilde{\bW}_1,\bx_{12})\\
        \vdots\\
        f(\tilde{\bW}_M,\bx_{Mn})
    \end{bmatrix} \in \mathbb{R}^N.
\end{align}
\label{defn:y_and_y_tilde}
\end{defn}

\begin{defn}(Proxy output and vector of proxy outputs at iteration $t$)

Given local models $\tilde{\bW}_1, \tilde{\bW}_2,\dots, \tilde{\bW}_M$, the proxy output $\tilde{f}(\bW,\bx_{ij})$ for any $\bW \in \mathbb{R}^d$ and for any $\bx_{ij}$ where $ i \in [M], j \in [n]$ is defined as,
\begin{align}
\tilde{f}(\bW,\bx_{ij}) = \phi(\tmi,\bx_{ij})^{\top}\bW.
\end{align}
\label{def:proxy_output}

Let $\bW^{(0)}, \bW^{(1)},\dots $ be the sequence of iterates generated by the global optimization process in FedFisher (\Cref{algo1} Lines 12-14). We define $\tft(t)$ as follows:
\begin{align}
    \tft(t) = \begin{bmatrix}
        \tilde{f}(\bW^{(t)},\bx_{11})\\
        \tilde{f}(\bW^{(t)},\bx_{12})\\
        \vdots\\
        \tilde{f}(\bW^{(t)},\bx_{Mn})
    \end{bmatrix} \in \mathbb{R}^N.
\label{def:vec_of_proxy_output}
\end{align}

\end{defn}

\begin{clm}(Bounded gradient)
For any $\bW \in \mathbb{R}^d$ and $\norm{\bx} \leq 1$ we have $\norm{\nabla_{\bW} f(\bW,\bx)}^2 \leq 1$.
\label{eq:phi_bound}
\end{clm}

\textbf{Proof.}

Firstly observe that,
\begin{align}
    \nabla_{\bW} f(\bW,\bx) = \phi(\bW,\bx).
\label{eq:grad_simplification}
\end{align}

We have,
\begin{align}
    \norm{\phi(\bW,\bx)}^2 &= \frac{1}{m}\sum_{r=1}^m\norm{a_r\bx \ind{\bx^{\top}\bw_r \geq 0}}^2 \nonumber\\
    & \leq \frac{1}{m}\sum_{r=1}^m \norm{a_r \bx}^2\nonumber \\
    & \leq 1 &&(\text{$a_r$'s are $\pm 1 $, $\norm{\bx} \leq 1$}).
\end{align}

\begin{clm}(Fisher closed-form expression) Under the squared loss $\ell(z,y) = \frac{1}{2}(y-z)^2$, the Fisher at client $i$ simplifies to,
\begin{align}
\bF_i &= \frac{1}{n}\sum_{j=1}^n \phi(\tmi,\bx_{ij}) \phi(\tmi,\bx_{ij})^{\top}.
\end{align}
\label{clm:fisher_claim}
\end{clm}

\textbf{Proof.}

Under the squared loss, we have $\prob(y|\bx_{ij},\bW) \propto \exp(-(y - f(\bW,\bx_{ij}))^2/2)$. Thus,

\begin{align}
    \nabla_{\bW} \log \prob(y|\bx_{ij},\bW)
    &= (f(\bW,\bx_{ij}) - y)\nabla_{\bW} f(\bW,\bx_{ij}) \nonumber \\
    & = (f(\bW,\bx_{ij}) - y)\phi(\bW,\bx_{ij}
) &&(\text{using \Cref{eq:grad_simplification}}).
\end{align}

Thus,
\begin{align}
    \bF_i &= \frac{1}{n}\sum_{j=1}^n \mathbb{E}_{y} \left[\nabla_{\bW} \log \prob(y|\bx_{ij},\bW) \nabla_{\bW} \log \prob(y|\bx_{ij},\bW)^{\top} \right]_{\bW = \tmi} \nonumber \\
    & =  \frac{1}{n}\sum_{j=1}^n \mathbb{E}_{y} \left[(y-f(\tmi,\bx_{ij})^2 \phi(\tmi,\bx_{ij})\phi(\tmi,\bx_{ij})^{\top} \right] \nonumber\\
    & = \frac{1}{n}\sum_{j=1}^n \mathbb{E}_{y} \left[(y-f(\tmi,\bx_{ij})^2 \right] \phi(\tmi,\bx_{ij}) \phi(\tmi,\bx_{ij})^{\top}  \nonumber\\
    & \overset{(a)}{=} \frac{1}{n}\sum_{j=1}^n \phi(\tmi,\bx_{ij}) \phi(\tmi,\bx_{ij})^{\top}
\end{align}
where $(a)$ uses $\prob (y|f(\tmi,\bx_{ij})) \propto \exp(-(y - f(\tmi,\bx_{ij}))^2/2)$ implying $y|f(\tmi,\bx_{ij}) \sim \mathcal{N}(f(\tmi,\bx_{ij}),1)$, following \Cref{assum:1}. 

\begin{clm}
\label{clm:min_eigen}
Define the matrix $\bH^{\infty}_i \in \mathbb{R}^{n \times n}$ as $(\bH_i^\infty)_{k,l} = \mathbb{E}_{\bw \sim \mathcal{N}(\bm{0},\bm{I})}\left[\bx_{ik}^{\top}\bx_{il}\mathbb{I}\left\{ \bw^{\top}\bx_{ik} \geq 0\right\}\mathbb{I}\left\{\bw^{\top}\bx_{il} \geq 0\right\} \right]$. Then $\lambda_{\min}(\bH^{\infty}_i) \geq \lambda_{\min} (\bH^\infty) = \lambda_0$.
\end{clm}

\textbf{Proof.} Suppose $\lambda_{\min}(\bH^{\infty}_i) = \lambda_i < \lambda_{\min} (\bH^\infty)$. Let $\bx$ be such that $\bx^{\top}\bH^{\infty}_i\bx = \lambda_i$. Define $\tilde{\bx} \in \mathbb{R}^{N \times N}$ to be a vector such that $\tilde{\bx}_{(i-1)\times n+1:i\times n} = \bx$ and zero everywhere else. Then $\tilde{\bx}^{\top}\bH^{\infty}\tilde{\bx} = \bx^{\top}\bH_i^\infty\bx = \lambda_i < \lambda_{\min} (\bH^\infty)$ leading to a contradiction. Therefore $\lambda_{\min}(\bH^{\infty}_i) \geq \lambda_{\min} (\bH^\infty) = \lambda_0$.

\begin{clm}
\label{clm:diag_inv}
Let $\bX_1,\bX_2\dots,\bX_M \in \mathbb{R}^{n \times p}$. Define,
\begin{align*}
    \bX = \begin{bmatrix}
        \bX_1 \\
        \bX_2\\
        \vdots\\
        \bX_M
    \end{bmatrix}
    \in \mathbb{R}^{N \times p}
\end{align*}
Also define 
\begin{align}
    \bD_{\bX} = \begin{bmatrix}
        \bX_1\bX_1^{\top} & \bm{0} & \cdots & \bm{0}\\
        \bm{0} & \bX_2\bX_2^{\top} & \cdots & \bm{0}\\
        \vdots\\
        \bm{0} & \bm{0} & \cdots & \bX_M\bX_M^{\top} \\
    \end{bmatrix} \in \mathbb{R}^{N \times N}
\end{align}

Then $ M\bD_{\bX} \succcurlyeq \bX\bX^{\top} $.
\end{clm}

\textbf{Proof.} Let $\ba = \text{vec}(\ba_1, \ba_2,\dots,\ba_M) \in \mathbb{R}^N$ where each $\ba_i \in \mathbb{R}^n$. Then,
\begin{align*}
    \ba^{\top}(M\bD_{\bX} - \bX\bX^{\top})\ba & = 
    M\sum_{i=1}^M \ba_i^{\top}\bX_i\bX_i^\top\ba_i - \sum_{i=1}^M\sum_{j=1}^M\ba_i^{\top}\bX_i\bX_j^{\top}\ba_j\\
    & = (M-1)\sum_{i=1}^M \ba_i\bX_i\bX_i^{\top}\ba_i - \sum_{i=1}^M\sum_{j=1, j\neq i}^M \ba_i^{\top}\bX_i\bX_j^{\top}\ba_j\\
    &= \sum_{i=1}^M\sum_{j=i+1}^M \ba_i^{\top}\bX_i\bX_i^{\top}\ba_i + \ba_j^{\top}\bX_j\bX_j^{\top}\ba_j - \ba_i^{\top}\bX_i\bX_j^{\top}\ba_j - \ba_j^{\top}\bX_j\bX_i^{\top}\ba_i\\
    & = \sum_{i=1}^M\sum_{j=i+1}^M\norm{\bX_i^{\top}\ba_i - \bX_j^{\top}\ba_j}^2\\
    & \geq 0
\end{align*}
Thus $M \bD_{\bX} \succcurlyeq \bX^{\top}\bX$.

\begin{clm}
\label{clm:actual_diag_inv}
\begin{align}
    M \begin{bmatrix}
        \bH_1^\infty & \bm{0} & \cdots & \bm{0}\\
        \bm{0} & \bH_2^{\infty} & \cdots & \bm{0}\\
        \vdots\\
        \bm{0} & \bm{0} & \cdots &  \bH_M^{\infty}\\
    \end{bmatrix} \succcurlyeq \bH^{\infty}
\end{align}
where $\bH_i^{\infty}$ is defined in \Cref{clm:min_eigen} and $\bH^{\infty}$ is defined in \Cref{defn:gram_matrix}. 
\end{clm}

\textbf{Proof.} This follows from an application of \Cref{clm:diag_inv}.

\subsubsection{Key Lemmas}

\label{subsec:bouning_training_loss}
Before moving to the theorem proof, we first state some key lemmas and their proofs. Note that the probabilities of all events in this proof are over the random initialization of $\bW_0$.

\Cref{lemma:local_opt} is used to bound the local optimization at clients. Note that since the local optimization at each client is independent of all other clients, we can consider them to be $M$ instances of centralized optimization starting from the same initialization. Thus we can use existing results in the centralized setting to give these bounds.

\begin{lem}(Theorem 3.1 and Lemma C.1 in \cite{arora2019fine})
If we set $m = \Omega \left( \frac{N^6}{\lambda_0^4\kappa^2\delta^3}\right)$ and $\eta = \bigO{\frac{\lambda_0}{N^2}}$, then with probability at least $1-\delta$ over the random initialization of $\bW_0$, for all clients $i \in [M]$ simultaneously, we have

\begin{enumerate}
\item $\sum_{j=1}^n (y_{ij} - f(\bW_0, \bx_{ij}))^2 = \bigO{\frac{N}{\delta}} $
\item $\sum_{j=1}^n (y_{ij} - f(\tmi,\bx_{ij}))^2 \leq (1-\eta \lambda_0/2)^{K}\bigO{\frac{N}{\delta}} $
\item $\norm{\wlmi - \wimi} \leq \frac{ 4\sqrt{N}(1-(1-\eta\lambda_0/4)^K)}{\sqrt{m}\lambda_0}\sqrt{\sum_{j=1}^n (y_{ij} - f(\bW_0, \bx_{ij}))^2} = R_0 \;  \forall r \in [m]$.
\end{enumerate}
where $\wlmi$ denotes the $r$-th weight in the local model of the $i$-th client and $\wimi$ denotes the $r$-th weight of the randomly initialized model $\bW_0$. 
\label{lemma:local_opt}
\end{lem}

Firstly note that the local optimization at client $i$ depends on $\bH_i^\infty$. Here we are using the result in \Cref{clm:min_eigen} which shows that $\lambda_{\min}(\bH_i^\infty) \geq \lambda_{\min} (\bH^{\infty}) = \lambda_0$. Also note that using \cite{arora2019fine} we can only guarantee that the above event holds at a single client with probability $1-\delta$. In order for this event to hold simultaneously with probability $1-\delta$ for all clients, we use the union bound and set the failure probability at each client to be $\delta' = \delta/M$. This leads the bound to have a dependence on $N$ instead of $n$ as in the single client case.

Also note that substituting the bound in part (1) in part (3), we have the following upper bound on $R_0$:
\begin{align}
\label{eq:R_0_upper_bound}
    R_0 &= \mathcal{O}\left(\frac{N(1-(1-\eta \lambda_0/4)^K)}{\sqrt{m \delta} \lambda_0} \right) \nonumber\\
    & = \mathcal{O}\left(\frac{N}{\sqrt{m \delta} \lambda_0} \right).
\end{align}

\Cref{lemma:A_ijr} and \Cref{lemma:du_lemma} are standard in two-layer NN optimization and are adopted as is.

\begin{lem}
\label{lemma:prob(A_ijr) bound} (Lemma 3.2 in \cite{du2018gradient})
For a given $R$, define the following event:
\begin{align}
    A_{ijr}(R) = \{\exists \bw: \norm{\bw - \wimi} \leq R, \ind{\bx_{ij}^{\top}\wimi \geq 0} \neq \ind{\bx_{ij}^{\top}\bw\geq 0}\}.
\end{align}
We have $\Pr(A_{ijr}(R)) \leq \frac{2R}{\sqrt{2\pi}\kappa}$ where the randomness is over the initialization of $\bw_{0,r}$.
\label{lemma:A_ijr}
\end{lem}

\begin{lem} (Lemma 3.1 in \cite{du2018gradient}). If $m = \Omega\left( \frac{N^2}{\lambda_0^2}\log \frac{N}{\delta}\right)$, we have with probability $1-\delta$, $\norm{\bH_{0} - \bH^\infty} \leq \frac{\lambda_0}{4}$  and $\lambda_{\min}(\bH_{0}) \geq \frac{3\lambda_0}{4}$.
\label{lemma:du_lemma}
\end{lem}

\Cref{lemma:eigen_value}, \Cref{lemma:global_opt} and \Cref{lem:orthogonal_neurons_size} are our contribution and form the basis of our proof.

\begin{lem}
If $m = \Omega \left( \frac{N^6}{\lambda_0^4\kappa^2\delta^3}\right)$, with probability $1-\delta$, $\lambda_{\min}(\tilde{\bH}) \geq \lambda_0/2$ and $\lambda_{\max}(\tilde{\bH}) \leq N$
\label{lemma:eigen_value}
\end{lem}

\textbf{Proof.}

The $(k,l)$-th entry of $\tilde{\bH} \in \mathbb{R}^{N \times N}$ is given by,
\begin{align}
   \tilde{\bH}_{kl} =  \frac{1}{m}\bx_{k'k''}^{\top}\bx_{l'l''} \sum_{r=1}^m \ind{\bx_{k'k''}^{\top}\tilde{\bw}_{k',r} \geq 0} \ind{\bx_{l'l''}^{\top}\tilde{\bw}_{l',r} \geq 0}
\end{align}

where $k' = \lceil k/n \rceil, k'' = k- k' + 1, l'= \lceil l/n \rceil, l''=l-l'+1$.

We have,
\begin{align}
    & \E{|\tilde{\bH}_{kl} - (\bH_{0})_{kl}|} \nonumber \\
    & = \E{\frac{1}{m}\left|\bx_{k'k''}^{\top}\bx_{l'l''}\right| \sum_{r=1}^m \ind{\ind{\bx_{k'k''}^{\top}\tilde{\bw}_{k',r} \geq 0} \ind{\bx_{l'l''}^{\top}\tilde{\bw}_{l',r} \geq 0} \neq \ind{\bx_{k'k''}^{\top}\wimi \geq 0} \ind{\bx_{l'l''}^{\top}\wimi \geq 0}}} \nonumber \\
    & \overset{(a)}{\leq} \E{\frac{1}{m}\sum_{r=1}^m \ind{\ind{\bx_{k'k''}^{\top}\tilde{\bw}_{k',r} \geq 0} \ind{\bx_{l'l''}^{\top}\tilde{\bw}_{l',r} \geq 0} \neq \ind{\bx_{k'k''}^{\top}\wimi \geq 0} \ind{\bx_{l'l''}^{\top}\wimi \geq 0}} }\nonumber \\
    & \leq \E{\frac{1}{m}\sum_{r=1}^m \ind{\ind{\bx_{k'k''}^{\top}\tilde{\bw}_{k',r} \geq 0}  \neq \ind{\bx_{k'k''}^{\top}\wimi \geq 0}} +\ind{\ind{\bx_{l'l''}^{\top}\tilde{\bw}_{l',r} \geq 0} \neq \ind{\bx_{l'l''}^{\top}\wimi \geq 0}}} \nonumber\\
    & \overset{(b)}{\leq} \E{\frac{1}{m}\sum_{r=1}^m \ind{A_{k'k''r}(R_0) } + \ind{A_{l'l''r}(R_0)}} \nonumber \\
    & = \frac{1}{m}\sum_{r=1}^m \Pr(A_{k'k''r}(R_0)) + \Pr(A_{l'l''r}(R_0)) \nonumber \\
    & \overset{(c)}{\leq} \frac{4R_0}{\sqrt{2\pi}\kappa}, \label{lem9.5:3}
\end{align}

where $(a)$ uses $\left|\bx_{k'k''}^{\top}\bx_{l'l''}\right| \leq \norm{\bx_{k'k''}}\norm{\bx_{l'l''}}\leq 1$ (\Cref{assum:data_normalization}), $(b)$ follows from the definition of $A_{ijr}$ in \Cref{lemma:A_ijr} and definition of $R_0$ in \Cref{lemma:local_opt}, $(c)$ uses the result in \Cref{lemma:A_ijr}.

Thus we have,
\begin{align}
    \E{\|\tilde{\bH}-\bH_{0}\|_{F}} \leq \E{\sum_{k,l} | \tilde{\bH}_{kl} - (\bH_{0})_{kl}|} \leq \frac{4N^2R_0}{\sqrt{2\pi}\kappa}.
\end{align}
By Markov's inequality, with probability $1-\delta$, we have
\begin{align}
   \|\tilde{\bH}-\bH_{0}\|_{F} \leq \frac{ \E{\|\tilde{\bH}-\bH_{0}\|_{F}}}{\delta} \leq \frac{4N^2R_0}{\sqrt{2\pi}\delta \kappa}.
\end{align}
Thus,
\begin{align}
    \norm{\tilde{\bH}-\bH_{0}} \leq  \|\tilde{\bH}-\bH_{0}\|_{F} \leq \frac{4N^2R_0}{\sqrt{2\pi}\delta\kappa}.
\label{eq:h_norm_bound}
\end{align}
This implies,
\begin{align}
    \lambda_{\min}(\tilde{\bH}) \geq \lambda_{\min}(\bH_{0}) - \frac{4N^2R_0}{\sqrt{2\pi}\delta\kappa} \geq \frac{\lambda_0}{2}
\end{align}
using \Cref{lemma:du_lemma} and substituting $m = \Omega\left(\frac{N^6}{\lambda_0^4\delta^3\kappa^2}\right)$ in the upper bound on $R_0$ in \Cref{eq:R_0_upper_bound}.

We also have,
\begin{align}
    \norm{\tilde{\bH}} &= \norm{\tilde{\bA}\tilde{\bA}^{\top}} \nonumber \\
    &= \norm{\tilde{\bA}^{\top}\tilde{\bA}} \nonumber \\
    & = \norm{\sum_{i=1}^M \sum_{j=1}^n\phi(\tmi,\bx_{ij})\phi(\tmi,\bx_{ij})^{\top}} \nonumber \\
    & \leq \sum_{i=1}^M \sum_{j=1}^n\norm{\phi(\tmi,\bx_{ij})\phi(\tmi,\bx_{ij})^{\top}} \nonumber \\
    & \leq \sum_{i=1}^M \sum_{j=1}^n\norm{\phi(\tmi,\bx_{ij})}^2 \nonumber \\
    & \leq N && (\text{\Cref{eq:phi_bound}})\label{lem9:5.4}.
\end{align}

Therefore $\lambda_{\max}(\tilde{\bH}) = \norm{\tilde{\bH}} \leq N$.
\qed

\begin{lem}
If we set $m = \Omega \left( \frac{N^6}{\lambda_0^4\kappa^2\delta^3}\right)$ and $\eta = \bigO{\frac{\lambda_0}{N^2}}$, then with probability at least $1-\delta$ over the random initialization of $\bW_0$, we have

\begin{enumerate}
\item $\norm{\tft(0)-\truey}^2 = \bigO{\frac{N^3}{\delta\lambda_0^2}}$

\item $\norm{\tft(t)-\truey}^2 \leq (1-\eta_S \lambda_0/2n)^t\norm{\tft(0)-\truey}^2$

\item $\norm{\womi - \wami} \leq \frac{ 4\sqrt{N}}{\sqrt{m}\lambda_0}\norm{\tft(0)-\truey} = R_1$.
\end{enumerate}
\label{lemma:global_opt}
\end{lem}

\textbf{Proof.}

Part $(1)$. 

We have,
\begin{align}
    (\tilde{f}(\bW^{(0)},\bx_{ij}) - \tilde{y}_{ij})^2 &= (\phi(\tmi,\bx_{ij})^{\top}\bW^{(0)} -\phi(\tmi,\bx_{ij})^{\top}\tmi)^2 \nonumber \\
    & = \left(\frac{1}{\sqrt{m}}\sum_{r=1}^m a_r \ind{\bx_{ij}^{\top}\wlmi \geq 0}\bx_{ij}^{\top}\left(\sum_{l=1}^M\tilde{\bw}_{l,r}/M - \wlmi\right) \right)^2 \nonumber \\
    & \leq  \sum_{r=1}^m \left( a_r \ind{\bx_{ij}^{\top}\wlmi \geq 0}\bx_{ij}^{\top}\left(\sum_{l=1}^M\tilde{\bw}_{l,r}/M - \wlmi\right) \right)^2 && (\text{Jensen's inequality}) \nonumber \\
    & \overset{(a)}{\leq} \sum_{r=1}^m \norm{\sum_{l=1}^M\tilde{\bw}_{l,r}/M - \wlmi}^2 \nonumber \\
    & \leq \frac{1}{M}\sum_{r=1}^m \sum_{l=1}^M \norm{\tilde{\bw}_{l,r} - \tilde{\bw}_{i,r}}^2 && (\text{Jensen's inequality}) \nonumber \\
    & \leq \frac{2}{M} \sum_{r=1}^m \sum_{l=1}^M  \norm{\tilde{\bw}_{l,r} - \bw_{0,r}}^2 + \norm{ \tilde{\bw}_{i,r}-\bw_{0,r}}^2 \nonumber\\
    & \leq 4mR_0^2 && (\text{\Cref{lemma:local_opt} Part 3}) \nonumber \\
    &= \bigO{\frac{N^2}{\lambda_0^2\delta}} && (\text{\Cref{eq:R_0_upper_bound}})\label{lem6.1.3}
\end{align}
where $(a)$ uses Cauchy-Schwartz and $\norm{a_r \ind{\bx_{ij}^{\top}\tilde{\bw}_{i,r} \geq 0}\bx_{ij}} \leq 1$.

Thus, $\norm{\tft(0)-\truey}^2 = \sum_{i=1}^M \sum_{j=1}^n  (\tilde{f}(\bW^{(0)},\bx_{ij}) - \tilde{y}_{ij})^2 = \bigO{\frac{N^3}{\lambda_0^2\delta}}$.

Part $(2)$.

The GD step that the central server performs for \ts{\algoname} can be written as,
\begin{align}
\label{eq:lemma_6_iterate}
    \bW^{(t+1)} &= \bW^{(t)} - \eta_S \sum_{i=1}^M \bF_i\left(\bW^{(t)} - \tmi \right) \nonumber \\
    & = \bW^{(t)} - \tilde{\eta_S} \sum_{i=1}^M\sum_{j=1}^n \phi(\tmi,\bx_{ij}) \phi(\tmi,\bx_{ij})^{\top}\left(\bW^{(t)} - \tmi \right) &&(\text{\Cref{clm:fisher_claim}, $\tilde{\eta_S} = \eta_S/n$}) \nonumber \\
     & = \bW^{(t)} - \tilde{\eta_S} \sum_{i=1}^M\sum_{j=1}^n \phi(\tmi,\bx_{ij}) \left(\tilde{f}(\bW^{(t)},\bx_{ij}) - \tilde{y}_{ij} \right) && (\text{\Cref{def:proxy_output}, \Cref{defn:y_and_y_tilde}})\nonumber \\
     & = \bW^{(t)} - \tilde{\eta_S} \tilde{\bA}^{\top}(\tft(t)-\truey) &&(\text{\Cref{defn:tilde_A}}).
\end{align}

 We have,
 \begin{align}
 \label{eq:server_opt_gd}
     \tft(t+1) - \tft(t) &= \tilde{\bA}(\bW^{(t+1)} - \bW^{(t)}) \nonumber \\
     & = -\tilde{\eta_S}\tilde{\bH}(\tft(t)-\truey) && (\text{\Cref{defn:tilde_A}, \Cref{eq:lemma_6_iterate}}).
 \end{align}

 Therefore,
 \begin{align}
     \norm{\tft(t+1) - \truey}^2 &= \norm{\tft(t+1) -  \tft(t) + \tft(t) - \truey}^2 \nonumber \\
     & = \norm{\tft(t) -  \truey}^2 - 2\tilde{\eta}_S(\tft(t) - \truey)\tilde{\bH}(\tft(t)-\truey) &&(\text{\Cref{eq:server_opt_gd}}) \nonumber\\
     &\quad + \tilde{\eta}_S^2\norm{\tilde{\bH}(\tft(t)-\truey)}^2 \nonumber \\
     & \leq (1-\tilde{\eta}_S \lambda_0 + \tilde{\eta}_S^2N^2) \norm{\tft(t) -  \truey}^2 && (\text{\Cref{lemma:eigen_value}}) \nonumber \\
     &\leq \left( 1-\frac{\eta_S \lambda_0}{2n}\right)\norm{\tft(t) -  \truey}^2 && \left(\tilde{\eta}_S \leq \frac{\lambda_0}{2N^2}\right). \label{eq:func_lemma_6}
 \end{align}

Part $(3)$.

From \Cref{eq:lemma_6_iterate} we have,

\begin{align}
\label{eq:lemma_6_1}
    \norm{\bw_{r}^{(t+1)} - \bw_{r}^{(t)}} 
    & \overset{(a)}{=} \norm{\tilde{\eta_S} \tilde{\bA}^{\top}_{(r-1)p+1:rp} (\truey - \tft(t))} && \nonumber \\
    & \leq \tilde{\eta}_S \norm{\tilde{\bA}^{\top}_{(r-1)p+1:rp} } \norm{\truey - \tft(t)} && (\text{Cauchy Schwartz}) \nonumber \\
    & \leq \tilde{\eta}_S \left\|\tilde{\bA}^{\top}_{(r-1)p+1:rp}\right\|_{F} \norm{\truey - \tft(t)} \nonumber \\
    &\leq \frac{\tilde{\eta}_S \sqrt{N}}{\sqrt{m}} \norm{\truey - \tft(t)} &&(\text{\Cref{defn:tilde_A}, $\norm{\bx_{ij}} \leq 1$})
\end{align}
where in $(a)$ we use the notation \text{$\bA_{x:y}$ to denote the submatrix of $\bA$ having row numbers $x$ to $y$}.

Therefore,

\begin{align}
     \norm{\womi - \bw_r^{(0)}} &\leq \sum_{t=0}^\infty \norm{\bw_{r}^{(t+1)} - \bw_{r}^{(t)}} \nonumber \\
     & \leq \frac{\tilde{\eta}_S \sqrt{N}}{\sqrt{m}} \sum_{t=0}^\infty \norm{\truey - \tft(t)} && (\text{using \Cref{eq:lemma_6_1}}) \nonumber \\
     & \leq \frac{\tilde{\eta}_S \sqrt{N}}{\sqrt{m}} \sum_{t=0}^\infty \left(1-\frac{\tilde{\eta}_S \lambda_0}{4}\right)^{t}\norm{\truey- \tft(0)} && (\text{using \Cref{eq:func_lemma_6}}) \nonumber\\
     & = \frac{4\sqrt{N}}{\sqrt{m}\lambda_0}\norm{\truey - \tft(0)}.
\end{align}

\qed

\begin{lem}
\label{lemma:S_ij bound}
Let $S_{ij} = \{r \in [m]: \ind{\bx_{ij}^{\top}\womi \geq 0} = \ind{\bx_{ij}^{\top}\wlmi \geq 0}\}\}$ and $S_{ij}^{\perp} = [m] \backslash S_{ij}$.
With probability $1-\delta$ over the initialization, we have $\sum_{i=1}^M\sum_{j=1}^n|S_{ij}^{\perp}|^2 = \bigO{\frac{m^2N^2(R_0^2 + R_1^2)}{\delta^2\kappa^2}}$ where $R_0$ is defined in \Cref{lemma:local_opt} and $R_1$ is defined in \Cref{lemma:global_opt}.
\label{lem:orthogonal_neurons_size}
\end{lem}

\textbf{Proof}.

We have,
\begin{align}
    \E{|S_{ij}^{\perp}|} &= \sum_{r=1}^m \Pr(\ind{\bx_{ij}^{\top}\womi \geq 0} \neq \ind{\bx_{ij}^{\top}\wlmi \geq 0}) \nonumber \\
    &   \leq \sum_{r=1}^m \Pr\left(\left\{\ind{\bx_{ij}^{\top}\womi\geq 0} \neq \ind{\bx_{ij}^{\top}\bw_{0,r} \geq 0}\right\} \cup \left\{\ind{\bx_{ij}^{\top}\wlmi \geq 0} \neq \ind{\bx_{ij}^{\top}\bw_{0,r} \geq 0}\right\}\right) \nonumber\\
    & \overset{(a)}{\leq}  \sum_{r=1}^m  \Pr\left(\left\{\ind{\bx_{ij}^{\top}\womi\geq 0} \neq \ind{\bx_{ij}^{\top}\bw_{0,r} \geq 0}\right\}\right) + \Pr\left(\left\{\ind{\bx_{ij}^{\top}\wlmi \geq 0} \neq \ind{\bx_{ij}^{\top}\bw_{0,r} \geq 0}\right\}\right) \nonumber\\
    &\overset{(b)}{\leq}  \sum_{r=1}^m   \Pr(A_{ijr}(R_0 + R_1)) + \Pr(A_{ijr}(R_0)) \nonumber \\
    & \overset{(c)}{\leq} \frac{4m(R_0+R_1)}{\sqrt{2\pi}\kappa} \label{lemma3:2}
\end{align}
where $(a)$ uses union bound, $(b)$ uses 
\Cref{lemma:prob(A_ijr) bound} and $\norm{\bw_r^* - \bw_{0,r}} \leq \norm{\bw_r^* - \bw_r^{(0)}} + \norm{\bw_r^{(0)}-\bw_{0,r}} = \norm{\bw_r^* - \bw_r^{(0)}} + \norm{\frac{1}{M}\sum_{i=1}^M \wlmi - \bw_{0,r}} \leq \norm{\bw_r^* - \bw_r^{(0)}} + \frac{1}{M}\sum_{i=1}^M \norm{\wlmi-\bw_{0,r}} \leq R_1 + R_0$ and $\norm{\wlmi-\bw_{0,r}} \leq R_0$, $(c)$ again uses \Cref{lemma:A_ijr}.

Thus, using Markov's inequality, with probability at least $1-\delta$, we have for any $i \in [M]$ and $j \in [n]$,
\begin{align}
\label{eq:single_S_ij_bound}
    |S_{ij}^{\perp}| 
 = \bigO{\frac{m(R_0 + R_1)}{\delta\kappa}}.
\end{align}

Setting the failure probability as $\delta/N$ in \Cref{eq:single_S_ij_bound}, we have with probability $1-\delta$, for all $i \in [M]$ and $j \in [n]$ simultaneously,

\begin{align}
     \forall i \in [M]; \forall j \in [n]: |S_{ij}^{\perp}| = 
 \bigO{\frac{mN(R_0 + R_1)}{\delta\kappa}}.
\end{align}

This implies,
\begin{align}
\sum_{i=1}^M\sum_{j=1}^n|S_{ij}^{\perp}|^2 & = \bigO{\frac{m^2N^2(R_0^2 + R_1^2)}{\delta^2\kappa^2}}.
\end{align}
\qed

\subsubsection{Proof of \Cref{theorem:fisher_avg_error}}

We first state the full theorem statement with the exact dependence of $m$ on $(N,\lambda_0^{-1},\delta^{-1},\kappa^{-1})$.

\begin{thm}
Under Assumptions \ref{assum:data_normalization}, \ref{assum:data_separation}, \ref{assum:local_training}, for $m =   \Omega \left( \frac{N^9}{\lambda_0^8\delta^4\kappa^2}\right)$, and i.i.d Gaussian initialization weights of $\bW_0$ as $\bw_{0,r} \sim \mathcal{N}(\bm{0},\kappa)$, and initializing the second layer weights $a_r = \{-1,1\}$ with probability $1/2$ for all $r \in [m]$, for step sizes $\eta = \mathcal{O}(\lambda_0/N^2)$, $\eta_S = \mathcal{O}(\lambda_0/N^2)$ and for a given failure probability $\delta \in (0,1)$, the following is true with probability $1-\delta$ over the random initialization:
\begin{align}
   L(\bW^*) \leq  \underbrace{\mathcal{O}\left((1-\eta\lambda_0/2)^K \frac{N}{\delta}\right)}_{\textnormal{local optimization error}} + \underbrace{\bigO{(2 - (1-\eta \lambda_0/2)^K)\frac{N^9}{\lambda_0^8\delta^4m}}}_{\textnormal{Laplace approximation error}}  .
\end{align}
\end{thm}

\textbf{Proof.}

Note that the conditions on $m$, $\eta$ and $\eta_S$ in \Cref{lemma:local_opt}, \Cref{lemma:du_lemma}, \Cref{lemma:eigen_value} and \Cref{lemma:global_opt} are satisfied by setting $m =   \Omega \left( \frac{N^9}{\lambda_0^8\delta^4\kappa^2}\right)$, $\eta = \mathcal{O}(\lambda_0/N^2)$, $\eta_S = \mathcal{O}(\lambda_0/N^2)$ and hence we can now apply these lemma results for our proof. Furthermore, we can scale down the failure probability in these lemmas by a constant factor to ensure that all the results in the lemmas hold simultaneously with high probability via union bound. 

First observe that setting $t \rightarrow \infty$ in \Cref{lemma:global_opt} part (2), we have,

\begin{align}
\label{obs:1}
    \tilde{f}(\bW^*,\bx_{ij}) = \tilde{y}_{ij}
\end{align}

Now,

\begin{align}
\label{eq:error_bound}
    L(\omi) &=\frac{1}{N}\sum_{i=1}^M\sum_{j=1}^n (f(\omi,\bx_{ij})-y_{ij})^2 \nonumber \\
    &= \frac{1}{N}\sum_{i=1}^M \sum_{j=1}^n\left((\phi(\omi,\bx_{ij})^{\top}\bW^* - \tilde{y}_{ij} + \tilde{y}_{ij} - y_{ij}\right)^2 && (\text{\Cref{eq:alternate_output_exp}}) \nonumber \\
    & \leq \underbrace{\frac{2}{N}\sum_{i=1}^M \sum_{j=1}^n \left(\phi(\omi,\bx_{ij})^{\top}\omi - \tilde{y}_{ij} \right)^2}_{T_1} + \underbrace{\frac{2}{N}\sum_{i=1}^M\sum_{j=1}^n (\tilde{y}_{ij} - y_{ij})^2}_{T_2}.
\end{align}

$T_2$ measures how well the local models fit their local data and can be bounded as 
\begin{align}
\label{eq:T_1_bound}
    T_2 = \mathcal{O}\left((1-\eta\lambda_0/2)^K \frac{N}{\delta}\right)
\end{align}
using the result from \Cref{lemma:local_opt}.

We now bound $T_1$ as follows:

\begin{align}
    T_1 &= \frac{2}{N}\sum_{i=1}^M \sum_{j=1}^n \left(\phi(\omi,\bx_{ij})^{\top}\omi - \tilde{y}_{ij} \right)^2 \nonumber \\
     &= \frac{2}{N}\sum_{i=1}^M \sum_{j=1}^n \left(\phi(\omi,\bx_{ij})^{\top}\omi - \tilde{f}(\bW^*,\bx_{ij}) \right)^2 && (\text{\Cref{obs:1}}) \nonumber \\
    &= \frac{2}{N}\sum_{i=1}^M \sum_{j=1}^n \left(\phi(\omi,\bx_{ij})^{\top}\omi - \phi(\tmi,\bx_{ij})^{\top}\omi\right)^2 && (\text{\Cref{def:proxy_output}}) \nonumber \\
    & = \frac{2}{N}\sum_{i=1}^M\sum_{j=1}^n \left(\frac{1}{\sqrt{m}}\sum_{r=1}^m a_r \bx_{ij}^{\top}\womi\left(\ind{\bx_{ij}^{\top}\womi \geq 0} - \ind{\bx_{ij}^{\top}\wlmi \geq 0}\right)\right)^2 \nonumber \\
     & \overset{(a)}{=} \frac{2}{N}\sum_{i=1}^M\sum_{j=1}^n\left(\frac{1}{\sqrt{m}}\sum_{r \in S_{ij}^{\perp}} a_r \bx_{ij}^{\top}\womi\left(\ind{\bx_{ij}^{\top}\womi \geq 0} - \ind{\bx_{ij}^{\top}\wlmi \geq 0}\right)\right)^2 \nonumber\\
     & \leq \frac{2}{N}\sum_{i=1}^M\sum_{j=1}^n\frac{|S_{ij}^{\perp}|}{m}\sum_{r \in S_{ij}^{\perp}}\left( a_r \bx_{ij}^{\top}\womi\right)^2 && (\text{Jensen's inequality})\nonumber\\
      & \overset{(b)}{\leq} \frac{2}{N}\sum_{i=1}^M \sum_{j=1}^n \frac{|S_{ij}^{\perp}|}{m}\sum_{r \in S_{ij}^{\perp}}\left(\bx_{ij}^{\top}\womi - \bx_{ij}^{\top}\wlmi\right)^2 \nonumber \\
     & \leq \frac{2}{N}\sum_{i=1}^M \sum_{j=1}^n\frac{|S_{ij}^{\perp}|^2}{m} \max_{r \in [m]} \norm{\womi-\wlmi}^2 \nonumber \\
       & \overset{(c)}{=} \bigO{\frac{m^2N(R_0^4+R_1^4)}{\delta^2 \kappa^2 m}} \nonumber\\
     & \overset{(d)}{=} \bigO{\frac{N^9}{\lambda_0^8\delta^4\kappa^2m}(2 - (1-\eta \lambda_0/2)^K)} \label{eq:T_2_bound}
\end{align}

where $(a)$ follows from the definition of $S_{ij}^{\perp}$ in \Cref{lem:orthogonal_neurons_size}. $(b)$ uses the observation that since $r \in S_{ij}^\perp$ we have $\text{sign}(\bx_{ij}^{\top}\womi) \neq \text{sign}(\bx_{ij}^\top\wlmi)$ which implies $|\bx_{ij}^{\top}\womi| \leq |\bx_{ij}^{\top}\womi - \bx_{ij}^\top\wlmi|$. For $(c)$ we use \Cref{lem:orthogonal_neurons_size} to bound $\sum_{i=1}^M \sum_{j=1}^n |S_{ij}|^2$ as $\bigO{\frac{m^2N^2(R_0^2 + R_1^2)}{\delta^2\kappa^2}}$ and $\norm{\bw_r^* - \tilde{\bw}_{i,r}}^2 \leq 2\norm{\bw_r^* - \bw_{r}^{(0)}}^2 + 2\norm{\bw_r^{(0)} - \tilde{\bw}_{i,r}}^2 \leq 2R_0^2 + 2R_1^2$. For $(d)$ we use $R_0 = \bigO{\frac{N}{\sqrt{m\delta}\lambda_0}(1-(1-\eta \lambda_0/4)^K)}$ from \text{\Cref{lemma:local_opt}} and $R_1 = \bigO{\frac{N^2}{\sqrt{m\delta}\lambda_0^2 }}$ from \text{\Cref{lemma:global_opt}}.

Now substituting the bounds in \Cref{eq:T_1_bound} and \Cref{eq:T_2_bound} in \Cref{eq:error_bound}, we have,

\begin{align}
     L(\bW^*) \leq  \mathcal{O}\left((1-\eta\lambda_0/2)^K \frac{N}{\delta}\right) + \bigO{(2 - (1-\eta \lambda_0/2)^K)\frac{N^9}{\lambda_0^8\delta^4\kappa^2 m}}
\end{align}
which completes the proof.
\qed

\newpage

\subsection{Generalization Guarantees}

In this subsection we provide generalization guarantees for the \ts{\algoname} global model. To do so, we first introduce some additional assumptions and definitions.

\subsubsection{Additional Assumptions and Definitions}

\begin{defn}
A distribution $\xi$ over $\mathbb{R}^d \times \mathbb{R}$ is $(\lambda_0,\delta,N)$-non degenerate if for $N$ i.i.d samples $\{(\bx_k,y_k)\}_{k=1}^N$ from $\mathcal{D}$, with probability at least $1-\delta$, we have $\lambda_{\min}(\bH^\infty) \geq \lambda_0 > 0$. 
\end{defn}

\begin{assum}
\label{assum:data_distr}
Let $\bar{\xi}$ be the distribution from which the test data points are sampled. $\bar{\xi}$ is a $(\lambda_0, \delta/3, N)$-non-degenerate distribution and the data samples $\{(\bx_k,y_k)\}_{k=1}^N \in \mathcal{D}$ are i.i.d samples from $\bar{\xi}$. 
\end{assum}
\begin{rem}
Note that in \Cref{assum:data_distr}, we are only assuming that the collection of local data across clients is drawn $i.i.d$ from some distribution $\bar{\xi}$. This does not imply that the data at any particular client $i$ is drawn from $\bar{\xi}$, i.e., $\mathcal{D}_i \sim \bar{\xi}$ or that the data at client $i$ is $i.i.d$ with the data at client $j$ for $i \neq j$. The data in $\mathcal{D}$ can be partitioned arbitrarily across clients and we make no assumptions on this. For instance, consider the case with $M = 2$ clients, $\mathcal{D} = \{(x_k,y_k) \sim \mathcal{N}(0,\bI) \}_{k=1}^N$, $\mathcal{D}_1 = \{(x,y) \in \mathcal{D} \text{ such that } y \geq 0\}$ and $\mathcal{D}_2 = \{(x,y) \in \mathcal{D} \text{ such that } y <0\}$. Then clearly the data in $\mathcal{D}_1$ and $\mathcal{D}_2$ are not sampled from $\mathcal{N}(0,\mathbf{I})$ and are not $i.i.d$ with each other.  
\end{rem}

Given the test data distribution $\bar{\xi}$, for a function $f: \mathbb{R}^d \rightarrow \mathbb{R}$, we have the following definitions of population loss $L_{\bar{\xi}}(f)$ and empirical loss $L(f)$:
\begin{align}
    L_{\bar{\xi}}(f) = \Eg{(\bx,y) \sim \bar{\xi}}{\frac{1}{2}(f(\bx)-y)^2},
\end{align}
\begin{align}
    L(f) = \frac{1}{2N}\sum_{k=1}^N (f(\bx_k)-y_k)^2.
\end{align}

\begin{defn}(Rademacher complexity)
Given $N$ samples, $\bx_1,\bx_2,\dots,\bx_N$, the Rademacher complexity of a function class $\mathcal{F}$ that maps from $\mathbb{R}^d$ to $\mathbb{R}$ is defined as,
\begin{align}
    \mathcal{R}_S(\mathcal{F}) = \frac{1}{N}\Eg{\epsilon_1,\epsilon_2,\dots,\epsilon_N}{\sup_{f \in \mathcal{F}}\sum_{k=1}^N \epsilon_k f(\bx_k)}
\end{align}
where each $\epsilon_i$ is sampled independently from the Rademacher distribution $\mathrm{unif}(\{-1,1\})$.
\end{defn}

\subsubsection{Key Lemmas}

We now state some key lemmas that will be used in the generalization proof.

We begin with the following lemma from \cite{arora2019fine} which bounds the Rademacher complexity of the class of two-layer neural network functions with bounded
distance from initialization.

\begin{lem}(Lemma 5.4 in \cite{arora2019fine})
Given $R > 0$, with probability at least $1-\delta$ over the random initialization of 
($\bW_0,\ba)$, simultaneously for every $B > 0$, the function class
\begin{align}
\mathcal{F}_{R,B}^{\bW_0,\ba} = \{f_{\bW,\ba}: \norm{\bw_r - \bw_{0,r}} \leq R \; \forall r \in [m], \norm{\bW - \bW_0} \leq B\} \nonumber
\end{align}
has Rademacher complexity bounded as,
\begin{align}
    \mathcal{R}_{S}( \mathcal{F}_{R,B}^{\bW_0,\ba}) \leq \frac{B}{\sqrt{2N}}\left( 1 + \left(\frac{2 \log 2/\delta}{m}\right)^{1/4}\right) + \frac{2R^2\sqrt{m}}{\kappa} + R \sqrt{2\log \frac{2}{\delta}}.
\end{align}
\label{lemma:bounded_rademacher_nn}
\end{lem}

\Cref{lemma:rademacher_bound} is standard in generalization theory and used to bound the population loss in terms of the empirical loss and Rademacher complexity of the class of functions to which our estimator belongs.

\begin{lem}(\cite{mohri2018foundations}) Suppose $\frac{1}{2}(f(\bx)-y)^2$ is bounded in the range $[0,c]$. With probability $1-\delta$ over the random sampling of $\mathcal{D}$ we have, 
\begin{align}
    \sup_{f \in \mathcal{F}} \{L_{\bar{\xi}}(f) - L(f)\} \leq 2\mathcal{R}_S(\mathcal{F}) +3c\sqrt{\frac{\log 2/\delta}{2N}}. 
\end{align}
\label{lemma:rademacher_bound}
\end{lem}

Similar to \Cref{lemma:local_opt}, we can use existing results in the centralized setting to bound the total distance moved by the local model of a client from its initialization.

\begin{lem}(Lemma C.3 and Lemma 5.3 in \cite{arora2019fine}) If we set $m \geq \kappa^{-2}\mathrm{poly}(N, \delta^{-1}, \lambda_0^{-1})$ and $\eta = \bigO{\frac{\lambda_0}{N^2}}$, then with probability at least $1-\delta$ over the random initialization, we have

\begin{enumerate}
    \item
    $\norm{\bH_{0} - \bH^{\infty}} \leq \bigO{\frac{N\sqrt{\log N/\delta} }{\sqrt{m}}}$
    \item $\norm{\tmi - \bW_0} \leq \sqrt{\by_i^{\top}(\bH_i^\infty)^{-1}\by_i} + \mathcal{O}\left(\frac{N\kappa}{\lambda_0 \delta}\right) + \frac{\mathrm{poly}(N,\lambda_0^{-1},\delta^{-1})}{m^{1/4} \kappa^{1/2}} \hspace{20pt} \forall i \in [M]$
\end{enumerate}
where $\by_i = \mathrm{vec}(y_{i1},y_{i2},\dots, y_{in})$ and $\bH_i^{\infty}$ is defined in \Cref{clm:min_eigen}.
\label{lemma:local_gen_bound}
\end{lem}

Note that in order to ensure that part (2) holds simultaneously for all $i \in [M]$ we set the failure probability to be $\delta/M$ in Lemma 5.3 in \cite{arora2019fine}.

The following lemmas are our contribution and used to bound the distance of the \ts{FedAvg} model from initialization and the \ts{\algoname} model from the \ts{FedAvg} model respectively.

\begin{lem} (Bounding distance of \ts{FedAvg} model from initialization)

Assuming the conditions in \Cref{lemma:local_gen_bound} hold true we have,
\begin{align}
 \norm{\bW^{(0)} - \bW_0} \leq \sqrt{2\by^{\top}(\bH^{\infty})^{-1}\by} + \mathcal{O}\left(\frac{N\kappa}{\lambda_0\delta}\right) + \frac{\mathrm{poly}(N,\lambda_0^{-1},\delta^{-1})}{m^{1/4} \kappa^{1/2}}   
\end{align}
\label{lemma:global_gen_bound}
\end{lem}

\textbf{Proof.}
We have,
\begin{align}
    \norm{\bW^{(0)} - \bW_0}^2 & = \norm{\sum_{i=1}^M \tmi/M - \bW_0}^2 \nonumber\\
    & \leq \frac{1}{M}\sum_{i=1}^M \norm{\tmi - \bW_0}^2 \nonumber \\
    & \leq \left(  \frac{1}{M} \sum_{i=1}^M 2\by_i^{\top}(\bH_i^\infty)^{-1}\by_i\right) + \mathcal{O}\left(\frac{N^2\kappa^2}{\lambda_0^2 \delta^2}\right) + \frac{\mathrm{poly}(N,\lambda_0^{-1},\delta^{-1})}{m^{1/2} \kappa} && (\text{using \Cref{lemma:local_gen_bound}}) \nonumber\\
    & = 2\by^\top(M \mathrm{diag}(\bH_1^{\infty}, \bH_2^{\infty},\dots, \bH_M^{\infty}))^{-1}\by + \mathcal{O}\left(\frac{N^2\kappa^2}{\lambda_0^2 \delta^2}\right) + \frac{\mathrm{poly}(N,\lambda_0^{-1},\delta^{-1})}{m^{1/2} \kappa} \nonumber \\
    & \leq 2\by^{\top}(\bH^{\infty})^{-1}\by + \mathcal{O}\left(\frac{N^2\kappa^2}{\lambda_0^2 \delta^2}\right) + \frac{\mathrm{poly}(N,\lambda_0^{-1},\delta^{-1})}{m^{1/2} \kappa} &&(\text{using \Cref{clm:actual_diag_inv}})
\end{align}

Thus we have,
\begin{align}
    \norm{\bW^{(0)} - \bW_0} \leq \sqrt{2\by^{\top}(\bH^{\infty})^{-1}\by} + \mathcal{O}\left(\frac{N\kappa}{\lambda_0\delta}\right) + \frac{\mathrm{poly}(N,\lambda_0^{-1},\delta^{-1})}{m^{1/4} \kappa^{1/2}}  && (\sqrt{a+b} \leq \sqrt{a} + \sqrt{b})
\end{align}

This completes the proof.
\qed

\begin{lem} (Bounding distance of \ts{\algoname} model from \ts{FedAvg} model)
If we set $m \geq \kappa^{-2}\mathrm{poly}(N, \delta^{-1}, \lambda_0^{-1})$, $\eta = \bigO{\frac{\lambda_0}{N^2}}$ and $\kappa = \mathcal{O}\left( \frac{\lambda_0 \delta}{N}\right)$ then with probability at least $1-\delta$ over the random initialization, we have,
\begin{align}
    \norm{\omi-\bW^{(0)}} = \mathcal{O}\left( \frac{N}{\lambda_0}\right)
\end{align}

\end{lem}

\textbf{Proof.}

We have,
\begin{align}
    \norm{\omi-\bW^{(0)}}^2 &= \norm{\sum_{t=0}^{\infty} \bW^{(t+1)} - \bW^{(t)}}^2 \nonumber \\
    &= \norm{-\etas \sum_{t=0}^{\infty} \tilde{\bA}^{\top}(\tft(t) - \truey) }^2 &&(\text{\Cref{eq:lemma_6_iterate}})\nonumber \\
    &= \norm{\etas  \sum_{t=0}^{\infty} \tilde{\bA}^{\top}(\bI - \tilde{\eta}_S\tilde{\bH})^t(\tft(0) - \truey)}^2 && (\text{\Cref{eq:server_opt_gd}}) \nonumber \\
    & = \norm{\tilde{\bA}^{\top}\bT(\tft(0) - \truey)}^2 && \left(\bT := \tilde{\eta}_S 
    \sum_{t=0}^\infty (\bI - \tilde{\eta}_S\tilde{\bH})^t\right) \nonumber \\
    & = (\tft(0) - \truey)^{\top}\bT\tilde{\bH}\bT(\tft(0) - \truey) && (\tilde{\bH} = \tilde{\bA}\tilde{\bA}^{\top})\nonumber \nonumber \\
    & \overset{(a)}{\leq}\frac{2\norm{\tft(0) - \truey}^2}{\lambda_0} \nonumber \\
    & \overset{(b)}{=} \mathcal{O}\left(\frac{N^2}{\lambda_0^2}\right).
\end{align}

For $(a)$ we use the following argument. Let $\tilde{\bH} = \bV\bD\bV^{\top}$ be the eigen decomposition of $\tilde{\bH}$. We see that $\bT = \tilde{\eta}_S\bV \sum_{t=0}^\infty (\bI - \tilde{\eta}_S \bD)\bV^{\top} = \bV\bD^{-1}\bV^{\top}.$ Thus $\bT\tilde{\bH}\bT = \bV\bD^{-1}\bV^{\top} = \tilde{\bH}^{-1}$. Furthermore $\norm{\tilde{\bH}^{-1}} \leq \frac{2}{\lambda_0}$ which follows from \Cref{lemma:eigen_value}. 

For $(b)$ we use the following argument,
\begin{align}
    \norm{\tft(0) - \truey}^2 &= \sum_{i=1}^M\sum_{j=1}^n (\tilde{f}(\bW^{(0)},\bx_{ij}) - \tilde{y}_{ij})^2 \nonumber\\
    &= \sum_{i=1}^M\sum_{j=1}^n (\phi(\tmi,\bx_{ij})^{\top}\bW^{(0)} -\phi(\tmi,\bx_{ij})^{\top}\tmi)^2 \nonumber \\
    & \leq \sum_{i=1}^M\sum_{j=1}^n \norm{\phi(\tmi,\bx_{ij})}^2 \norm{\bW^{(0)} - \tmi}^2 &&(\text{Cauchy-Schwartz}) \nonumber\\
    & \leq \sum_{i=1}^M\sum_{j=1}^n \norm{\bW^{(0)} - \tmi}^2 && (\text{\Cref{eq:grad_simplification}}) \nonumber \\
    & \leq 2\sum_{i=1}^M\sum_{j=1}^n \norm{\bW^{(0)} - \bW_0}^2 + \norm{\tmi - \bW_0}^2 \nonumber\\
    & \overset{(c)}{=} \sum_{i=1}^M \sum_{j=1}^n \mathcal{O}\left(\frac{N}{\lambda_0} \right) \nonumber \\
    & = \mathcal{O}\left(\frac{N^2}{\lambda_0} \right). \label{eq:global_init_bound}
\end{align}

where $(c)$ follows from \Cref{lemma:local_gen_bound} part (2), \Cref{lemma:global_gen_bound} and setting $\kappa = \mathcal{O}\left( \frac{\lambda_0 \delta}{N}\right)$.

Thus,
\begin{align*}
  \norm{\omi-\bW^{(0)}} = \mathcal{O}\left(\frac{N}{\lambda_0} \right). 
\end{align*}
\qed

\subsubsection{Generalization Theorem and Proof}

We first state the generalization bound and then provide the proof below.

\begin{thm}
\label{theorem:fisher_avg_generalization}
Fix a failure probability $\delta \in (0,1)$. Under Assumptions \ref{assum:data_normalization}, \ref{assum:data_separation}, \ref{assum:data_distr}, if we let $\kappa = \mathcal{O}\left(\frac{\lambda_0 \delta}{N} \right)$, $m \geq  \kappa^{-2}\mathrm{poly}(N, \lambda_0^{-1}, \delta^{-1})$, $K = \Omega\left(\frac{1}{\eta \lambda_0}\log (N/\delta)\right)$  then with probability at least $1-\delta$ over the random initialization and sampling of training data points, we have
\begin{align}
    \mathbb{E}_{(\bx,y) \sim \bar{\xi}}[(f(\bW^*,\bx)-y)^2] \leq \mathcal{O}\left( \frac{\sqrt{N}}{\lambda_0}\right) + \sqrt{\frac{\by^{\top}(\bH^\infty)^{-1}\by}{N}} + \mathcal{O}\left(\sqrt{\frac{\log \frac{N}{\lambda_0 \delta}}{N}} \right)
\end{align}
\end{thm}

\begin{rem}
We see that the last two terms in our bound match the generalization bound of a model trained in the centralized setting on $\mathcal{D}$ \citep{arora2019fine}. The first term of order $\mathcal{O}\left( \frac{\sqrt{N}}{\lambda_0}\right)$ comes from bounding the distance of the \ts{\algoname} model from \ts{FedAvg}, i.e., $\norm{\bW^* - \bW^{(0)}}$ and can be seen as the additional error incurred in the federated setting by \ts{\algoname}. In particular, to bound $\norm{\bW^* - \bW^{(0)}}$ we need to bound $\norm{\tft(0) - \tilde{\by}}$ (see \Cref{lemma:global_gen_bound}). Clearly, if the data across clients is similar then $\norm{\tft(0) - \tilde{\by}}$ will be small. However, our current approach does not make any bounded heterogeneity assumptions leading to a pessimistic $\mathcal{O}\left( \frac{\sqrt{N}}{\lambda_0}\right)$ bound.
We conjecture that this bound can be improved by explicitly incorporating bounded data heterogeneity assumptions; however we leave the investigation of this as future work.
\end{rem}
\textbf{Proof.}

As a first step, we bound the distance of the $\omi$ from $\bW_0$. We have, 
\begin{align}
     \norm{\omi - \bW_0} &\leq \norm{\omi - \bW^{(0)}} + \norm{\bW^{(0)} - \bW_0} \nonumber \\
     &\leq \mathcal{O}\left(\frac{N}{\lambda_0} \right) +\sqrt{2\by^{\top}(\bH^{\infty})^{-1}\by} + \mathcal{O}\left(\frac{N\kappa}{\lambda_0\delta}\right) + \frac{\mathrm{poly}(N,\lambda_0^{-1},\delta^{-1})}{m^{1/4} \kappa^{1/2}} && (\text{\Cref{lemma:local_gen_bound}, \Cref{lemma:global_gen_bound}}) \nonumber\\
     & = \mathcal{O}\left(\frac{N}{\lambda_0} \right) \label{eq:total_dist_upper_bound}
\end{align}
where the last line follows from setting $\kappa = \mathcal{O}\left(\frac{\lambda_0 \delta}{N} \right)$, $m \geq  \kappa^{-2}\mathrm{poly}(N, \lambda_0^{-1}, \delta^{-1})$ and observing that $\sqrt{2\by^{\top}(\bH^{\infty})^{-1}\by} = \mathcal{O}(\sqrt{N/\lambda_0})$.

Next we aim to bound $\norm{\womi - \bw_{0,r}}$ for any $r \in [m]$. Recall we already bounded this quantity in \Cref{eq:T_2_bound} as follows,
\begin{align}
    \norm{\womi - \bw_{0,r}} &\leq \norm{\womi - \bw_r^{(0)}} + \norm{\bw_r^{(0)} - \bw_{0,r}} \nonumber \\
    & \leq \norm{\womi - \bw_r^{(0)}} + \frac{1}{M}\sum_{i=1}^M\norm{\tilde{\bw}_{i,r} - \bw_{0,r}} \nonumber \\
    & = R_1 + R_0 \nonumber &&(\text{\Cref{lemma:local_opt}, \Cref{lemma:global_opt}}) \\
    & =  \bigO{\frac{N^2}{\sqrt{m\delta}\lambda_0^2 }} + \bigO{\frac{N}{\sqrt{m\delta}\lambda_0}(1-(1-\eta \lambda_0/4)^K)} \nonumber \\
    & = \mathcal{O}\left( \frac{\mathrm{poly}(N,\delta^{-1},\lambda_0^{-1})}{m^{1/2}}\right) \label{eq:weight_dist_upper_bound}
\end{align}

We now use a similar argument as the proof of theorem 5.1 in \cite{arora2019fine} to provide the generalization guarantee. Under \Cref{assum:data_distr} we have $\mathcal{D} \sim \bar{\xi}$, where $\bar{\xi}$ is a $(\lambda_0, \delta/3,N)$ non-degenerate distribution. This implies with probability $1-\delta/3$ we have $\lambda_{\min}(\bH^\infty) \geq \lambda_0 > 0$. Assuming this event holds, we note that the following events hold with probability $1-\delta$ over the random initialization.

i)
\begin{align}
    L(\bW^*) \leq \frac{1}{\sqrt{N}} \label{eq:empirical_bound}
\end{align}
which follows from \Cref{theorem:fisher_avg_error} by setting $m = \Omega \left(\mathrm{poly}(N,\lambda_0^{-1},\delta^{-1},\kappa^{-1}) \right)$, $K = \Omega \left( \frac{1}{\eta \lambda_0} \log (N/\delta)\right)$,
$\eta = \mathcal{O}\left(\frac{\lambda_0}{N^2}\right)$ and $\eta_S = \mathcal{O}\left(\frac{\lambda_0}{N^2} \right)$.

ii) 
\begin{align}
    \norm{\omi - \bW_0} \leq B \text{ and } \norm{\womi - \bw_{0,r}} \leq R \text{ where } B  = \mathcal{O}\left(\frac{N}{\lambda_0} \right) \text{ and } R  = \mathcal{O}\left( \frac{\mathrm{poly}(N,\delta^{-1},\lambda_0^{-1})}{m^{1/2}}\right) \label{eq:R_B_bound}
\end{align}
which follows from \Cref{eq:total_dist_upper_bound} and \Cref{eq:weight_dist_upper_bound} respectively.

iii) For $k = 1,2,\dots,$ let $B_k = k$. Let $k^*$ be the smallest integer such that $B_{k^*} = k^* \geq B$. This implies $B_{k^*} \leq B+1$ and $k^* = \bigO{N/\lambda_0}$. Using \Cref{lemma:bounded_rademacher_nn} we have,
\begin{align}
   \mathcal{R}_{S}( \mathcal{F}_{R,B_{k^*}}^{\bW_0,\ba}) &\leq \frac{B + 1}{\sqrt{2N}}\left( 1 + \left(\frac{2 \log 2/\delta}{m}\right)^{1/4}\right) + \frac{2R^2\sqrt{m}}{\kappa} + R \sqrt{2\log \frac{2}{\delta}} \nonumber \\
   & \leq  \mathcal{O}\left( \frac{\sqrt{N}}{\lambda_0}\right) + \sqrt{\frac{\by^{\top}(\bH^\infty)^{-1}\by}{N}} + \frac{2}{\sqrt{N}} \label{eq: rade_bound}
\end{align}
where the last inequality follows from setting $m \geq \mathrm{poly}(N,\lambda_0^{-1},\delta^{-1},\kappa^{-1})$, $\kappa = \bigO{\frac{\lambda\delta}{N}}$ and \Cref{eq:total_dist_upper_bound}. Also note that $f_{\bW^*, \ba} \in \mathcal{F}_{R,B_k^*}^{\bW_0,\ba}$.

iv) Using \Cref{lemma:rademacher_bound} and a union bound over all $k \in \{0,1,2,\mathcal{O}(N/\delta)$\} we have,
\begin{align}
    \sup_{f \in \mathcal{F}_{R,B_k}^{\bW_0,\ba}} \{L_{\bar{\xi}}(f) - L(f)\} \leq 2 \mathcal{R}_{S}(\mathcal{F}_{R,B_k}^{\bW_0,\ba}) + \bigO{\sqrt{\frac{\log (N/\delta \lambda_0)}{N}}}\;\; \forall k \in \{0,1,2,\mathcal{O}(N/\delta)\}. \label{eq:gen_rade_bound}
\end{align}

Note that we can scale down the failure probabilities in each of the above events to ensure that all the above events hold simultaneously with probability $1-\delta$, via union bound. Now assuming all the above conditions hold simultaneously, we have,
\begin{align}
    L_{\bar{\xi}} (f_{\bW^*, \ba}) &\leq L(f_{\bW^*,\ba}) + 2 \mathcal{R}_{S}(\mathcal{F}_{R,B_{k^*}}^{\bW_0,\ba}) + \mathcal{O}\left(\sqrt{\frac{\log \frac{N}{\lambda_0 \delta}}{N}} \right) && (\text{from \Cref{eq:gen_rade_bound}}) \nonumber \\
    &\leq \frac{1}{\sqrt{N}}+ 2 \mathcal{R}_{S}(\mathcal{F}_{R,B_{k^*}}^{\bW_0,\ba}) + \mathcal{O}\left(\sqrt{\frac{\log \frac{N}{\lambda_0 \delta}}{N}} \right) && (\text{from \Cref{eq:empirical_bound}}) \nonumber \\
    & \leq \mathcal{O}\left( \frac{\sqrt{N}}{\lambda_0}\right) + \sqrt{\frac{\by^{\top}(\bH^\infty)^{-1}\by}{N}} + \mathcal{O}\left(\sqrt{\frac{\log \frac{N}{\lambda_0 \delta}}{N}} \right) && (\text{from \Cref{eq:R_B_bound} and \Cref{eq: rade_bound}}) \nonumber \\
\end{align}
which completes the proof.
\qed

\newpage

\section{Communication Efficiency of \ts{\algoname} Variants}
\label{sec:appendix_comm_eff}
We assume that the number of bits used to represent a scalar is $32$ by default.
For \ts{FedAvg}, clients just need to transfer $\tmi$, making the total communication cost $32d$ bits. Our goal in this section is to show that we can introduce compression techniques in \ts{\algonamed} and \ts{\algonamek} to match the communication cost of \ts{FedAvg} while having similar accuracy as the uncompressed version of these algorithms. To do so, we use standard uniform quantization and SVD compression as described below.

\paragraph{Uniform Quantization.}  Let $s_q \in \{1,2,\dots,16\}$ be the factor by which we want to compress our information. We define number of quantization levels as $ l_q = 2^{\floor{32/s_q}-1}-1$. Now given a vector $\bx \in \mathbb{R}^d$, we define each element of the quantized $\bx$ as follows,
\begin{align}
    [Q(\bx,s_q)]_i = \|\bx\|_{\infty}\text{sign}(x_i)\zeta_i(\bx,s_q)
\end{align}
where $\|\bx\|_{\infty} = \max_{i \in [d]} |x_i|$ and $\zeta_i(\bx,s_q) = \ceil[\Big]{l_q\frac{|x_i|}{\|\bx\|_{\infty}}}/l_q$. Note that $\zeta_i(\bx,s_q)$ can take only $l_q + 1 = 2^{\floor{32/s_q}-1}$ distinct values and therefore to communicate $\zeta_i(\bx,s_q)$ we only need $\floor{32/s_q}-1$ bits. To communicate $\text{sign}(x_i)$ we need $1$ bit and to communicate $\|\bx\|_{\infty}$ we need $32$ bits. Thus the communication cost of $Q(\bx,s_q)$ becomes $d(\floor{32/s_q}-1) + d +32 = d(\floor{32/s_q}) + 32$ bits.

\paragraph{Singular Value Decomposition Compression.} Let $\bA = \mathbb{R}^{(m \times m)}$ matrix. The SVD decomposition of $\bA$ can be written as,
\begin{align}
    \bA = \bU\bSigma\bV^{\top}
\end{align}
where $\bU = \mathbb{R}^{(m \times m)}$ is the matrix of left singular vectors, $\bSigma \in \mathbb{R}^{(m \times m)}$ is a diagonal matrix with each element corresponding to a singular value and $\bV = \mathbb{R}^{(m \times m)}$ is the matrix of right of singular vectors. The singular values are assumed to be sorted by magnitude, i.e., $|\Sigma_{1,1}| \geq |\Sigma_{2,2}|\dots, |\Sigma_{m,m}|$. We see that the total cost for communicating $\bA$ will $32m^2$ bits. To reduce this cost, a standard idea is to send only a limited number of singular values and singular vectors obtained by the SVD decomposition of $\bA$. Specifically let $s_v$ be the factor by which we want to compress the information in $\bA$. We define $l_v = \floor{m/2s_v}$. Then the SVD decompression of $\bA$ can be defined as,
\begin{align}
    V(\bA,s_v) = \bU_{l_v}\bSigma_{l_v}\bV_{l_v}^{\top}
\end{align}
where $\bU_{l_v} \in \mathbb{R}^{(m \times l_v)}$ corresponds to first $l_v$ columns of $\bU$, $\bSigma_{l_v} \in \mathbb{R}^{(l_v \times l_v)}$ is a diagonal matrix corresponding to the first $l_v$ elements of $\bSigma$ and $\bV \in \mathbb{R}^{(m \times l_v)}$ corresponds to the first $l_v$ columns of $\bV$. The communication cost of $V(\bA,s_v)$ becomes $32(ml_v + l_v + ml_v) \approx 64ml_v \leq 32m^2/s_v$ bits, thereby achieving close to $s_v$ compression.

\paragraph{Compression in \ts{\algonamed}.} For \ts{\algonamed}
, clients need to communicate $\tmi$ and $\afi$ where the number of parameters in $\afi$ is exactly $d$. To ensure comparable communication to \ts{FedAvg}, we quantize the weights corresponding to each layer of a neural network in $\tmi$ and $\afi$ by a factor of 2, i.e, $s_q = 2$. This ensures that the communication of \ts{\algonamed} is $32(d + 2L)$ bits where $L$ is the number of layers in the neural network. We note that there is a small overhead of $64L$ bits; however is negligible since $d \gg 2L$ for our neural networks. \Cref{table:fedfisherdiag_compression} shows the effects of compression on the accuracy of $\ts{\algonamed}$ for CIFAR-10 and FashionMNIST datasets. For FashionMNIST the compressed version even slightly improves upon the uncompressed version which can be attributed to an additional regularization effect of compression. For CIFAR-10, the drop in accuracy is less than $1\%$.

\begin{table}[h]
\centering
\caption{Test accuracy performance of \ts{\algonamek} with different levels of $s_q$ compression on FashionMNIST and CIFAR10 with 5 clients and $\alpha = 0.1$.}
\color{black}

\begin{tabular}{l|c|c|c}
\toprule
\multicolumn{1}{l}{Dataset} &
\multicolumn{1}{c}{$s_q$} &
\multicolumn{1}{c}{Compression} &
\multicolumn{1}{c}{Test Accuracy}\\
\midrule
 FashionMNIST & $1$ & $-$ & $54.64$\small{$\pm 4.91$}\\
 & $2$ & $2\times$ & $56.56$\small{$\pm 5.57$}\\
\midrule
 CIFAR-10 & $1$ & $-$ & $39.68$\small{$\pm 2.45$}\\
 & $2$ & $2\times $ & $39.44$\small{$\pm 2.31$}\\

\bottomrule
\end{tabular}
\label{table:fedfisherdiag_compression}
\end{table}

\paragraph{Compression in \ts{\algonamek}.} 
Let $\{m_0, m_1,m_2,\dots,m_L\}$ be the dimensions of each layer of a $L$ layer neural network with $m_0$ corresponding to the dimension of the input. 
For \ts{\algonamek}, $\afi$ can be represented as $\{(\bA_1 \otimes \bB_1),(\bA_2 \otimes \bB_2),\dots,(\bA_L \otimes \bB_L)\}$ where $\bA_l \in \mathbb{R}^{(m_{l-1}\times m_{l-1})}$ and $\bB_l \in \mathbb{R}^{(m_{l}\times m_{l})}$ represents the Kronecker factors of the $l$-th layer. Thus, the communication cost of $\afi$ in this case is $\sum_{l=1}^L 32(m_{l-1}^2 + m_l^2)$ bits. 

Our goal is to ensure that the communication cost of compressed $\afi$ is less than $16d$ bits (we compress $\tmi$ to $16d$ bits using quantization, similar to \ts{\algonamed}). To do so, we use a mix of quantization and SVD compression. Specifically each $\bA_l$ and $\bB_l$ is first compressed to the SVD decomposition corresponding to maintaining the top $l_v$ vectors. This ensures that the communication cost of $(\bA_l,\bB_l)$ is $32(2m_{l-1}l_v + 2m_ll_v + 2l_v)$. This SVD decomposition is then further compressed using $s_q$ quantization compression to ensure that the communication cost is $(32/s_q)(2m_{l-1}l_v + 2m_ll_v + 2l_v)$. We set $s_q$ and $l_v$ such that $\sum_{l=1}^L (32/s_q)(2m_{l-1}l_v + 2m_ll_v + 2l_v) \leq 16d$. The corresponding $s_v$ is then defined as $\ceil{m/2l_v}$. \Cref{table:fedfisherkfac_compression} summarizes the results obtained by different levels of $s_q$ and $s_v$ for the FashionMNIST and CIFAR datasets respectively. We see that keeping $s_q=4$ and setting $s_v$ accordingly usually gives the best performance and hence we use this setting for all our experiments.

\begin{table}[h]
\centering
\caption{Test accuracy performance of \ts{\algonamek} with different levels of $s_q \times s_v$ compression on FashionMNIST and CIFAR10 with 5 clients and $\alpha = 0.1$.}
\color{black}

\begin{tabular}{l|cc|c|c}
\toprule
\multicolumn{1}{l}{Dataset} &
\multicolumn{1}{c}{$s_q$} &
\multicolumn{1}{c}{$s_v$} &
\multicolumn{1}{c}{Compression} &
\multicolumn{1}{c}{Test Accuracy}\\
\midrule
 & $1$ & $1$ & $-$ & $68.47$\small{$\pm 2.73$}\\
 FashionMNIST & $2$ & $3$ & $6\times$ & $65.87$\small{$\pm 3.42$}\\
 & $4$ & $1.5$ & $6\times$ & $68.96$\small{$\pm 2.71$}\\
 & $6$ & $1$ & $6\times$ & $63.90$\small{$\pm 4.34$}\\
\midrule
& $1$ & $1$ & $-$ & $49.13$\small{$\pm 1.25$}\\
 CIFAR-10 & $2$ & $4$ & $8\times$ & $45.29$\small{$\pm 1.30$}\\
 & $4$ & $2$ & $8\times$ & $47.60$\small{$\pm 0.84$}\\
& $6$ & $1.33$ & $8\times$ & $43.53$\small{$\pm 1.85$}\\

\bottomrule
\end{tabular}
\label{table:fedfisherkfac_compression}
\end{table}

\newpage

\section{Additional Experimental and Details}
\label{sec:appendix_addnl_expts}
\subsection{Details on Synthetic Experiment}
Our setup consists of $M = 2$ clients and $n = 100$ data points at each client with $\bx \in \mathbb{R}^2$ and $y \in \mathbb{R}$. The data at each client is generated following a similar procedure as \cite{li2020federated}. For each $i \in [M]$, we first sample $w_i \sim \mathcal{N}(0,1)$, $b_i \sim \mathcal{N}(0,1)$, $\bw_i \sim \mathcal{N}(w_i, \bI) \in \mathbb{R}^2, \bb_i \sim \mathcal{N}(b_i, \bI) \in \mathbb{R}^2$. Then for all $j \in [n]$ we have $\tilde{\bx}_{ij} \sim \mathcal{N}(\bb_i, \bSigma)$, $\bx_{ij} = \tilde{\bx}_{ij}/\norm{\tilde{\bx}_{ij}}$ and $y_{ij} = \bw_{i}^{\top}\bx_{ij}$ where $\bSigma = \mathrm{diag}(1, 2^{-1.2})$.

For our two layer neural network, we initialize the weights in the second layer as $a_r = \frac{1}{\sqrt{m}}$ with probability $1/2$ or $a_r = -\frac{1}{\sqrt{m}}$ otherwise for all $r \in [m]$. We keep the weights in the second layer to be fixed as assumed in our analysis. The weights in the first layer are initialized as $\bw_r \sim \mathcal{N}\left(0,\frac{1}{2}\right)$ for all $r \in [m]$. We set $\eta = 0.1$ for the local optimization and $\eta_S = 0.001$ for the global optimization. Results in Figure $1(a)$ were averaged over $50$ seeds and results in Figure $1(b)$ were averaged over $10$ random seeds. 

\subsection{Experiment on Varying Number of Clients}

\begin{figure*}[hbt!]
    \centering
\includegraphics[width=1.00\columnwidth]{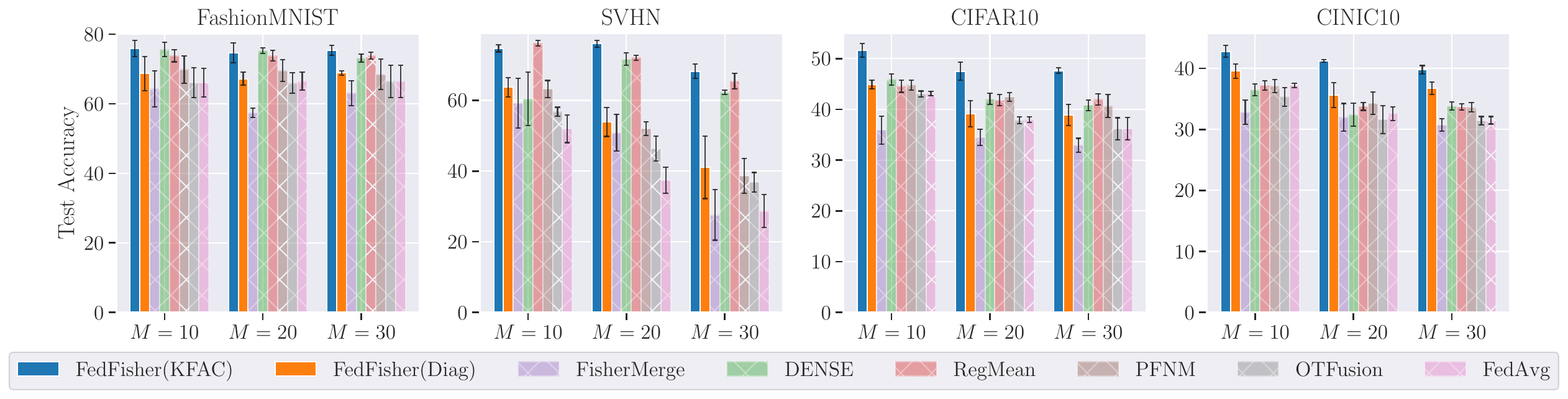}
    \caption{Test accuracy results on different datasets by keeping $\alpha = 0.3$ fixed and varying number of clients $M$. \ts{\algoname} variants, especially \ts{\algonamek}, consistently outperforms other baselines.}
    \label{fig:client_figure}
\end{figure*}

\Cref{fig:client_figure} shows the result our experiment where we keep the heterogeneity to be fixed as $\alpha = 0.3$ and vary the number of clients $M = \{10,20,30\}$. Note that as the total dataset size is fixed, as we increase $M$, each client gets assigned fewer data samples. Thus as $M$ increases, local models have a larger tendency to overfit the data they are trained on, making it harder to aggregate such models to achieve a global model with good performance. Nonetheless we see that \ts{\algoname} variants, especially \ts{\algonamek}, continue to outperform baselines even for large $M$ with up to $8\%$ improvement in some cases such as CIFAR-10.  

\subsection{Hyperparameter Details}

For hyperparameter tuning we assume that the server has access to a dataset of $500$ samples, sampled uniformly from the original training set. We describe the hyperparameters tuned for each of the algorithms below.

\paragraph{\ts{\algonamed} and \ts{\algonamek}.} While Algorithm 1 performs the server optimization with GD, this can be replaced with any other suitable GD optimizer. We choose to use the \ts{Adam} optimizer here. We set $\eta_S = 0.01, \beta_1 = 0.9, \beta_2 = 0.99$ and $\epsilon = 0.01$ for the \ts{Adam} optimizer  and number of steps $T = 2000$ for all our experiments. We measure the validation performance after every $100$ steps and use the model which achieves the best validation performance as the final \ts{FedFisher} model.

\paragraph{\ts{PFNM.}} For \ts{PFNM} we use the version applicable for CNNs whose code is available at \url{https://github.com/IBM/FedMA}. We keep the default values of the hyperparameters $\sigma = 1, \sigma_0 = 1$ and $\gamma = 7$ as in the original code in all our experiments.

\paragraph{\ts{DENSE}.} For \ts{DENSE} we use the default settings as described in Section 3.1.4 of \cite{zhang2022dense} and official code implementation available at \url{https://github.com/zj-jayzhang/DENSE}. In particular we set $T_G =30, \lambda_1 = 1, \lambda_2=0.5$ and train the server model with SGD optimizer with learning rate $\eta_S = 0.01$ and momentum factor $0.9$. We use the validation data to determine the model which achieves the best validation performance during the server training and use this as the final model. 

\paragraph{\ts{OTFusion}.} The official code for \ts{OTFusion} available at \url{https://github.com/sidak/otfusion} contains more than $15$ hyperparameters, making it hard to tune each of these parameters. Among these we found that the \ts{correction} and \ts{past-correction} hyperparameter affect performance most and hence we focus on tuning these hyperparameters for our experiments. In particular we vary \ts{correction} $\in \{\text{True}, \text{False}\}$ and \ts{past-correction} $\in \{\text{True}, \text{False}\}$ in our experiments and choose the model which achieves the best validation performance.

\paragraph{\ts{RegMean}.} For \ts{RegMean}, we keep all parameters the same as the official code available at \url{https://github.com/bloomberg/dataless-model-merging} and only tune $\alpha$ in the range $\{0.1, 0.9\}$ using validation data as suggested by the authors.

\paragraph{\ts{FisherMerge.}} For \ts{FisherMerge}, we compute the diagonal Fisher at each client using the entire dataset available at each client with a batch size of $1$ and set $\lambda_i = p_i$ where $p_i$ is the proportion of data available at client $i$. We also keep the \ts{fisher-floor} variable to be $10^{-6}$ as in the official code implementation available at \url{https://github.com/mmatena/model_merging/tree/master/model_merging}.

\subsection{Computation Details}

Experiments in \Cref{table:heterogeneity}, \Cref{table:fedfisherdiag_compression}, \Cref{table:fedfisherkfac_compression} and \Cref{fig:client_figure} were averaged over 5 seeds and run on a NVIDIA TitanX GPU. Results in \Cref{fig:multi_round} and \Cref{table:pretrained} were averaged over 3 seeds and run on NVIDIA A100 GPU.

\newpage

\section{Measuring Privacy Using Inversion Attacks}
\label{sec:appendix_privacy_expt}
\paragraph{Measuring Privacy using Inversion Attacks.} As discussed in \Cref{sec:practical_algo}, for \ts{\algonamek}, while the local models can be securely aggregated, the server does need access to individual K-FAC information at clients to perform the global optimization. We argue, however, that this is more privacy-preserving than having access to the local models which is needed for knowledge distillation and neuron matching baselines. We demonstrate this empirically via the following inversion attack on the MNIST dataset \cite{lecun1998mnist}. 

We consider a setup where client $1$ has all the images corresponding to label $3$ along with $1000$ randomly sampled images. The remaining data is then split among $9$ clients with $\alpha = 0.1$. The objective at the server is to reconstruct the data at client $1$ corresponding to label $3$. To do so, we adopt the attack strategy outlined in \cite{zhang2020secret}, a popular model inversion attack using GANS. We assume that the server has access to auxiliary data of size $5000$ that is randomly sampled from the data belonging to clients $2$ to $10$. Note that this ensures that the server does not have any samples corresponding to label $3$. We now consider three cases $\textbf{(i)}$ server has access to the local model at client $1$ (\ts{DENSE}, \ts{OTFusion}, \ts{PFNM}) $\textbf{
(ii)}$ server has access to the aggregate of local models (\ts{FedAvg}) and $\textbf{(iii)}$ server has access to the aggregate of local model plus the K-FAC computed by client $1$ (\ts{\algonamek}). To modify the attack to include the K-FAC information, we add an additional loss term that penalizes the difference between the K-FAC on the generated data and the given K-FAC. In particular, we use the squared loss between the elements of the generated K-FAC and given K-FAC, summed over all elements.

\begin{figure}[!h]
  \centering
   \subfloat[]{\includegraphics[width=0.3\linewidth]{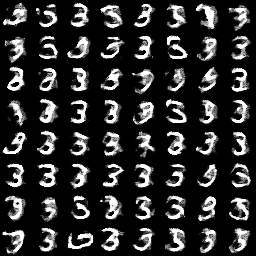}\label{fig:local_model_inv}}
   \hspace{50pt}
   \subfloat[]{\includegraphics[width=0.3\columnwidth]{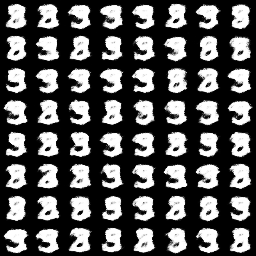}\label{fig:fisher_inv}}\\
   \vspace{-0.5em}
   \caption{Reconstructed images when (a) server has access to the local model at first client and (b) server has access to the global model and K-FAC information of first client. The goal is to generate images corresponding to digit $3$.}
  \label{fig:inv_attack}
\end{figure}

\Cref{table:inv_attack} summarizes the attack accuracy in the different cases. \Cref{fig:inv_attack} shows the generated images for the first and third case. We see that while adding the K-FAC information slightly improves the attack accuracy compared to case $\textbf{(ii)}$, the overall accuracy is still much lower than case \textbf{(i)} as also evidenced by the generated images. This hints at the difficulty of inverting the K-FAC to generate meaningful client data, especially when we do not have access to the local models themselves. We acknowledge that there could be more sophisticated attacks designed especially for inverting the K-FAC; however we leave a investigation of such methods to future work.

\begin{table}[h]
\centering
\caption{Inversion attack accuracy at server corresponding to different algorithms. While \ts{\algonamek} improves attack accuracy compared to \ts{FedAvg}, it is still lower than other one-shot baselines.}
\color{black}

\begin{tabular}{lc}
\toprule
Algorithm &
\multicolumn{1}{c}{Attack Accuracy}\\
\midrule
\ts{OTFusion}, \ts{PFNM}, \ts{DENSE}
& $78.4$\tiny{$\pm 3.0$}\\
\ts{FedAvg}
& \hspace{1pt} $5.0$\tiny{$\pm 1.0$}\\
\ts{\algonamek} & \hspace{1pt} $9.6$\tiny{$\pm 2.9$}\\
\bottomrule
\end{tabular}
\label{table:inv_attack}
\end{table}

\end{document}